\documentclass{article}

\usepackage[english]{babel} 		

\usepackage{amsfonts}				
\usepackage{amsmath}				
\usepackage{siunitx}  				
\DeclareSIUnit{\hours}{hours}
\usepackage{xfrac}					
\usepackage{bm}						
\usepackage{mathtools}
\usepackage{mathabx}				
\usepackage{nicefrac}

\usepackage{xspace}
\usepackage[utf8]{inputenc} 
\usepackage[T1]{fontenc}    
\usepackage{microtype}      
\usepackage{xcolor}         

\usepackage{hyphenat}
\usepackage[margin=1in]{geometry}
\usepackage[section]{placeins}
\usepackage{needspace}

\usepackage{graphicx}
\usepackage[labelfont=bf]{caption}
\usepackage{subcaption}
\newcommand{\leaderboardprofilefigure}[5]{%
    \begin{figure}[htbp]
        \centering
        \includegraphics[width=\textwidth]{#1}\par
        \begin{subfigure}[t]{0.39\textwidth}
            \captionsetup{justification=raggedright,singlelinecheck=false}
            \caption{Leaderboard}
            \label{#4}
        \end{subfigure}\hfill
        \begin{subfigure}[t]{0.52\textwidth}
            \captionsetup{justification=raggedright,singlelinecheck=false}
            \caption{Performance profiles}
            \label{#5}
        \end{subfigure}
        \caption{#2}
        \label{#3}
    \end{figure}%
}

\usepackage{tikz}
\usepackage{pifont}
\usepackage[export]{adjustbox}

\usepackage{enumitem}
\setlist{topsep=6pt plus 2pt minus 2pt,itemsep=2pt plus 1pt minus 1pt,parsep=0pt}

\usepackage{booktabs}               
\usepackage{multirow}               
\usepackage{soul}                   
\usepackage{changepage,threeparttable} 
\usepackage{tabularx}               
\usepackage{multirow}               
\usepackage{etoolbox}               
\usepackage{makecell}               

\usepackage{colortbl}               
\newcolumntype{g}{>{\columncolor{TUgray_vlight}}l}  
\newrobustcmd{\B}{\bfseries}

\definecolor{mydarkblue}{rgb}{0,0.08,0.45} 
\usepackage[
    colorlinks=true,
    linkcolor=mydarkblue,
    citecolor=mydarkblue,
    filecolor=mydarkblue,
    urlcolor=mydarkblue,
]{hyperref}       		
\usepackage{url}
\usepackage[nameinlink,capitalize,noabbrev]{cleveref}

\let\oldFootnote\footnote
\newcommand\nextToken\relax

\renewcommand\footnote[1]{%
    \oldFootnote{#1}\futurelet\nextToken\isFootnote}

\newcommand\isFootnote{%
    \ifx\footnote\nextToken\textsuperscript{,}\fi}

\usepackage{natbib}

\usepackage[textsize=footnotesize]{todonotes}
\setuptodonotes{bordercolor=white}

\usepackage[super]{nth}						                

\newcommand{\Runtime}{Runtime\xspace}

\newcommand{\wl}{workload\xspace}
\newcommand{\wls}{workloads\xspace}

\newcommand{\algoperf}{\textsc{AlgoPerf}\xspace}
\newcommand{\algoperfold}{\textsc{AlgoPerf}~v0.5\xspace}
\newcommand{\algoperfnew}{\textsc{AlgoPerf}~v1\xspace}

\newcommand{\sfadamwii}{\textsc{Schedule-Free AdamW v2 (PyTorch)}\xspace}
\newcommand{\ademamix}{\textsc{AdEMAMix (PyTorch)}\xspace}
\newcommand{\sfadamw}{\textsc{Schedule-Free AdamW (PyTorch)}\xspace}
\newcommand{\nadamwbase}{\textsc{NAdamW (Baseline v0.5) (JAX)}\xspace}
\newcommand{\sfadamwjaxii}{\textsc{Schedule-Free AdamW v2 (JAX)}\xspace}
\newcommand{\muonpt}{\textsc{Muon (PyTorch)}\xspace}
\newcommand{\sfadamwjax}{\textsc{Schedule-Free AdamW (JAX)}\xspace}
\newcommand{\nadamw}{\textsc{NAdamW (JAX)}\xspace}
\newcommand{\cautiousnadamw}{\textsc{Cautious NAdamW (JAX)}\xspace}
\newcommand{\muonjax}{\textsc{Muon (JAX)}\xspace}
\newcommand{\lion}{\textsc{Lion (PyTorch)}\xspace}

\newcommand{\dilocosw}{\textsc{Single-worker DiLoCo (JAX)}\xspace}

\newcommand{\ademamixadamw}{\textsc{AdEMAMix (AdamW equiv.) (PyTorch)}\xspace}

\newcommand{\imagenet}{\textsc{ImageNet}\xspace}
\newcommand{\wmt}{\textsc{WMT}\xspace}
\newcommand{\librispeech}{\textsc{LibriSpeech}\xspace}
\newcommand{\criteo}{\textsc{Criteo 1TB}\xspace}
\newcommand{\ogbg}{\textsc{OGBG}\xspace}
\newcommand{\fastmri}{\textsc{fastMRI}\xspace}

\newcommand{\finewebedu}{\textsc{FineWeb-Edu 10B}\xspace}

\newcommand{\resnetfifty}{\textsc{ResNet-50}\xspace}
\newcommand{\resnet}{\textsc{ResNet}\xspace}

\newcommand{\vit}{\textsc{ViT}\xspace}

\newcommand{\transformer}{\textsc{Transformer}\xspace}
\newcommand{\deepspeech}{\textsc{DeepSpeech}\xspace}
\newcommand{\conformer}{\textsc{Conformer}\xspace}
\newcommand{\dlrmsmall}{\textsc{DLRMsmall}\xspace}
\newcommand{\gnn}{\textsc{GNN}\xspace}
\newcommand{\unet}{\textsc{U-Net}\xspace}

\newcommand{\dolmii}{\textsc{\shortstack[l]{Decoder-only \\ Transformer}}\xspace}

\newcommand{\jax}{\textsc{JAX}\xspace}

\usepackage{amsmath,amsfonts,bm}

\def\eqref#1{equation~\ref{#1}}

\def\1{\bm{1}}

\DeclareMathAlphabet{\mathsfit}{\encodingdefault}{\sfdefault}{m}{sl}
\SetMathAlphabet{\mathsfit}{bold}{\encodingdefault}{\sfdefault}{bx}{n}

\definecolor{TUred}{RGB}{165,30,55}
\definecolor{TUdark}{RGB}{50,65,75}
\colorlet{TUdark_light}{TUdark!80}
\definecolor{TUgold}{RGB}{180,160,105}

\definecolor{TUgray}{RGB}{185,184,188}
\colorlet{TUgray_light}{TUgray!80}
\colorlet{TUgray_vlight}{TUgray!40}

\definecolor{TUdarkblue}{RGB}{65,90,140}
\definecolor{TUblue}{RGB}{0,105,170}
\definecolor{TUlightblue}{RGB}{80,170,200}

\definecolor{TUlightgreen}{RGB}{125,165,75}
\definecolor{TUgreen}{RGB}{125,165,75}
\definecolor{TUdarkgreen}{RGB}{50,110,30}

\definecolor{TUlightred}{RGB}{200,80,60}
\definecolor{TUpurple}{RGB}{175,110,150}
\definecolor{TUorange}{RGB}{210,150,0}

\definecolor{SNSblue}{rgb}{0.1216, 0.4666, 0.7059}
\definecolor{SNSorange}{rgb}{1.0, 0.4980, 0.0549}
\definecolor{SNSgreen}{rgb}{0.1725, 0.6274, 0.1725}
\definecolor{SNSred}{rgb}{0.84, 0.15, 0.16}
\definecolor{SNSpurple}{rgb}{0.58, 0.40, 0.74}

\definecolor{SNSorange_shaded}{HTML}{ffcea3}
\definecolor{SNSblue_shaded}{HTML}{8ebad9}
\definecolor{SNSgreen_shaded}{HTML}{cae7ca}
\definecolor{SNSred_shaded}{HTML}{ea9293}

\definecolor{MPLred_shaded}{HTML}{df735b}
\definecolor{MPLblue_shaded}{HTML}{3885bc}

\definecolor{PlotRed}{rgb}{0.77, 0.31, 0.32}
\definecolor{PlotBlue}{rgb}{0.30, 0.45, 0.69}

\definecolor{shampoosubcol}{rgb}{0.09, 0.745, 0.81}
\definecolor{sfadamcol}{rgb}{0.12, 0.467, 0.706}
\definecolor{generalizedadamcol}{rgb}{1.0, 0.5, 0.055}
\definecolor{cycliclrcol}{rgb}{0.1725, 0.6274, 0.1725}
\definecolor{nadampcol}{rgb}{0.8392, 0.1529, 0.1569}
\definecolor{baselinecol}{HTML}{32414b}
\definecolor{amoscol}{rgb}{0.737, 0.741, 0.133}
\definecolor{casprcol}{rgb}{0.549, 0.337, 0.294}
\definecolor{lawaqcol}{rgb}{0.58 0.40, 0.74}
\definecolor{lawaemacol}{rgb}{0.82, 0.73, 1.0}
\definecolor{sfprodigycol}{HTML}{696969}
\definecolor{nadamwseqcol}{rgb}{1.0, 0.498, 0.055}
\definecolor{sinvnumcol}{rgb}{0.58, 0.404, 0.74}
\definecolor{sinvcol}{rgb}{0.8392, 0.153, 0.157}
\definecolor{adamgcol}{rgb}{0.1725, 0.6274, 0.1725}			

\title{Towards Easier-to-Tune Training Algorithms: New Results from AlgoPerf}
\author{%
  Priya Kasimbeg$^{1}$ \quad
  Ahmed Khaled$^{1}$ \quad
  Aahlad Chandrabhatta$^{1}$ \quad
  Yifan Wang$^{1}$ \\[0.4ex]
  Andy Liu$^{2}$ \quad
  Niccol\`{o} Ajroldi$^{5}$ \quad
  David Tweedle$^{3}$ \quad
  Xingyu (Alice) Yang$^{6}$ \\[0.4ex]
  Isaac Patole \quad
  Frank Schneider$^{4}$ \quad
  George E. Dahl$^{1}$ \\[1.2ex]
  \small
  $^{1}$Google \quad
  $^{2}$AI For Tomorrow \quad
  $^{3}$Cohere\footnote{Work done while an independent researcher.} \quad
  $^{4}$University of T\"{u}bingen \quad
  $^{5}$Ellis Institute T\"{u}bingen \quad
  $^{6}$Goodfire\footnote{Work done while at Meta FAIR.}
}
\date{September 2026}

\begin{document}

\maketitle

\begin{abstract}
Evaluating deep learning optimizer is challenging because confounding factors such as hyperparameter tuning and optimizer-model co-design often obscure true algorithmic performance. We present results of the updated \algoperfnew benchmark, in which we evaluate a new set of training algorithms on wall-clock time-to-target across nine workloads on fixed hardware. This report focuses on the results of the benchmark's self-tuning ruleset, which requires submissions to use a single configuration across all workloads. Out of the 13 submissions in PyTorch and JAX, Schedule-Free AdamW ranked first. In addition to the wall-clock scores, we also analyze the trade-offs between step efficiency and wall-clock execution, revealing that matrix-preconditioned and two-loop methods achieve good sample efficiency but require more aggressive computational optimizations to be competitive in runtime. Finally, our hyperparameter saliency and retention analyses show that the learning rate and momentum are highly sensitive across domains and algorithms, highlighting the need for methods that are more robust against hyperparameter tuning and can generalize well to new tasks.
\end{abstract}

\section{Introduction}
Neural network training often relies on fragile empirical recipes governed by brittle hyperparameter tuning procedures and massive computational budgets. Even though training algorithms, optimizers, and their accompanying training recipes are such a crucial part of the neural network pipeline, it is very difficult to determine what the true state of the art is in training algorithms. Much of the literature on new optimizers suffers from confounding factors like weak baselines, optimizer-model co-design, limited testing across different tasks and obscured tuning recipes.

The \algoperf: Training Algorithms benchmark \citep{Dahl2023AlgoPerf} was designed to largely eliminate these confounding factors and measure practical speed-ups in neural network training achieved solely by improving the
underlying training algorithms. The benchmark was originally launched as a competition with two rulesets. In the \textit{external tuning} ruleset, submissions provided workload-agnostic hyperparameter search spaces and then were scored based on a sample from their hyperparameter space, while in the \textit{self-tuning} ruleset, the submissions were scored without any hyperparameter sampling. In both rulesets, submissions are compared on time-to-result across multiple deep learning workloads, training on fixed hardware. The results of the \algoperf competition \citep{Kasimbeg2025AlgoPerfResults} ranked the Distributed Shampoo \citep{shi2023distributeddataparallelpytorchimplementation} algorithm first on the external-tuning leaderboard, while Schedule-Free AdamW \citep{defazio2024roadscheduled} placed first on the self-tuning leaderboard. Since then, there has been a renewed interest in second-order methods like Muon \citep{jordan2024muon} and momentum-based or EMA methods like SNOO \citep{kallusky_snoo_step_nest25} and AdEMAMix \citep{pagliardini_the_adem_opti24}.

In this report we release a new set of results on an updated version of the \algoperf benchmark. Notably, compared to the original \algoperfold benchmark \citep{Dahl2023AlgoPerf}, the \algoperfnew benchmark introduced a new Decoder-only Language Modeling workload. \algoperfnew also upgraded the fixed hardware from an $8\times\text{V100}$ setup to a $4\times\text{A100}$ GPU configuration, and recalibrated step-time budgets for the hardware update.
We evaluated 13 self-tuning submissions for this report, which were implemented and submitted by the members of the MLCommons Algorithms Working Group and members of the external community. The results show an updated version of Schedule-Free AdamW ranking first, followed by \ademamixadamw (an AdEMAMix submission parameterized with a long slow-EMA warmup that renders it effectively equivalent to AdamW).

\section{Background}

\textbf{Benchmarks for language models.} The modded nanogpt challenge~\citep{modded_nanogpt_2024} benchmarks training speed on FineWeb-Edu~\citep{penedo_the_fine_data24}. However, submissions may simultaneously modify both the optimizer and the neural architecture, coupling architectural and algorithmic effects. Furthermore, its short training horizons can yield an incomplete picture of optimizer speedups over longer runs.

\citet{semenov_benc_opti_for25} and \citet{wen_fant_pret_opti25} benchmark optimizers for large language models under fixed architectures, exploring hyperparameter tuning, batch sizes, and training horizons. \citet{wen_fant_pret_opti25} find that optimizer rankings shift with the training horizon: Muon~\citep{jordan2024muon} excels on shorter runs, but is overtaken by SOAP~\citep{vyas_soap_impr_and24} and Kron~\citep{li_prec_stoc_grad15} on longer horizons. Under large batch sizes, \citet{wen_fant_pret_opti25} find that matrix-based optimizers consistently outperform scalar methods. Conversely, \citet{semenov_benc_opti_for25} observe that momentum-heavy, variance-reduced algorithms such as AdEMAMix~\citep{pagliardini_the_adem_opti24} and MARS~\citep{yuan_mars_unle_the24} scale particularly well to large batches. In \algoperf, submissions may tune the batch size per optimizer, while the language model architecture remains fixed.

\textbf{Benchmarks for other workloads.} \citet{schaipp_opti_benc_for25} benchmark optimizers for training diffusion models on dynamical systems, finding that matrix-preconditioned methods (Muon and SOAP) achieve substantially lower final losses than AdamW. In supervised tabular learning across 17 datasets, \citet{gorishniy_benc_opti_for26} evaluate held-out generalization rather than time to reach a specific loss, identifying Muon as the most consistent top performer, while noting that an Exponential Moving Average (EMA) of model weights provides a strong baseline over vanilla AdamW.

The evaluation by \citet{zhang_evol_of_opti26} is most closely related to ours, benchmarking 23 optimizers across vision (\resnetfifty, \vit-S) and language (Llama-60M) to assess learning rate sensitivity and architectural robustness. However, \algoperf's self-tuning track provides a much more stringent test of cross-domain generalization: it demands a single, fixed set of hyperparameters that succeeds across all 9 diverse deep learning workloads without any workload-specific tuning.

\section{Methodology}
We briefly summarize the \algoperf benchmark by explaining key terms. For a more detailed discussion of the benchmark’s motivation, description, and justification see the original benchmark paper \citep{Dahl2023AlgoPerf} and competition results paper \citep{Kasimbeg2025AlgoPerfResults}.
We made several modifications to the benchmark as it was initially proposed by \citet{Dahl2023AlgoPerf}. Specifically we:
\begin{enumerate}
\item Added a language modeling workload, consisting of a decoder-only NanoDO-style language model~\citep{nanodo} and the \finewebedu dataset.
\item Upgraded the fixed-hardware setup from an $8\times\text{V100}$ (16\,GB) to a $4\times\text{A100}$ (40\,GB) NVIDIA GPU setup.
\item Adjusted the runtime budgets based on the results of the \algoperfold competition and calibrated the runtime budgets further for the A100 hardware setup.
\item Migrated the \jax workloads from \texttt{pmap} sharding to \texttt{jit} sharding.
\item Extended the submission API to give more control over the batch normalization statistic updates.
\item Extended the submission API to specify the evaluation parameters at the beginning of an evaluation phase (useful for EMA methods). 
\end{enumerate}

A full list of changes can be found in the changelog in the \algoperf GitHub repository~\citep{AlgoPerfChangelog}.

\subsection{Workloads}
The benchmark features multiple neural network training tasks, called \textit{workloads}, each
consisting of a dataset, model, loss function, target metric, value for the \textit{validation target} and \textit{runtime budget}. The objective of each training algorithm
\textit{submission} is to minimize the wall-clock time needed to train a workload to reach the \textit{validation target} (time-to-target) within the \textit{runtime budget}.

The benchmark evaluates algorithms across the nine \algoperf base workloads covering multiple domains, as shown in \Cref{tab:workloads}.

\begin{table}[htbp]
\centering
\caption{Summary of the nine \algoperf base workloads, validation targets, and base runtime budgets. The self-tuning runtime budgets are $1.5\times$ these base values.}
\label{tab:workloads}
\begingroup
\footnotesize
\setlength{\tabcolsep}{4pt}
\setlength{\defaultaddspace}{2pt}
\noindent
\begin{tabularx}{\textwidth}{@{}XllllS[table-format=2.6]S[table-format=6.0]@{}}
	\toprule
	                                       &              &              &     &      & {Validation} & {Base \Runtime} \\
	\textbf{Task} & \textbf{Dataset} & \textbf{Model} & \textbf{Loss} & \textbf{Metric} & \textbf{Target} & \textbf{Budget (s)} \\ \midrule
	Clickthrough rate \newline prediction  & \criteo      & \dlrmsmall   & CE  & CE   & 0.123735 &  8915 \\ \addlinespace
	MRI reconstruction                     & \fastmri     & \unet        & L1  & SSIM & 0.723653 &  2745 \\ \addlinespace
	Image                                  & \imagenet    & \resnetfifty & CE  & ER   & 0.22569  & 49918 \\
	classification                         &              & \vit         & CE  & ER   & 0.22691  & 64292 \\ \addlinespace
	Speech                                 & \librispeech & \conformer   & CTC & WER  & 0.085884 & 43680 \\
	recognition                            &              & \deepspeech  & CTC & WER  & 0.119936 & 36949 \\ \addlinespace
	Molecular property \newline prediction & \ogbg        & \gnn         & CE  & mAP  & 0.28098  & 11303 \\ \addlinespace
	Language modeling                      & \finewebedu  & \dolmii    & CE  & PPL  & 22.2995  & 31967 \\ \addlinespace
	Translation                            & \wmt         & \transformer & CE  & BLEU & 30.8491  & 16114 \\ \bottomrule
\end{tabularx}
\par
\endgroup

\end{table}

\subsection{Submissions}
Training algorithm submissions must adhere to the fixed \algoperf API and are limited to five submission functions:
\begin{enumerate}
\item \texttt{update\_params} is responsible for modifying the network’s parameters during training and typically involves
optimization algorithms, such as SGD, ADAM, or a custom method.
\item \texttt{init\_optimizer\_state} initializes
the training algorithm’s internal state, e.g. to define learning rate schedules.
\item \texttt{data\_selection} controls how batches are constructed, to use techniques such as curriculum learning or data
echoing \citep{choi2019echoing}.
\item \texttt{get\_batch\_size} defines a batch size for each workload. This is the only workload-dependent parameter permitted by the API and allows participants to maximize hardware memory utilization on the fixed 4$\times$A100 GPU setup across models of different memory footprints.
\item \texttt{prepare\_for\_eval} prepares for evaluation, e.g., switching to evaluation parameters.
\end{enumerate}
This limited API isolates training speed-ups resulting from improvements to the training algorithm itself,
rather than pipeline optimizations or model changes. The API also
ensures that submissions can be applied to generic deep learning workloads by being fully specified
without any workload-specific behavior (apart from the batch size).

\subsection{Self-Tuning track}
The \algoperf benchmark distinguishes between the \emph{external-tuning} and \emph{self-tuning} evaluation tracks. The self-tuning track evaluates the out-of-the-box generalizability of optimization algorithms under strict, realistic deployment constraints.

\paragraph{Self-tuning ruleset vs.\ external-tuning.}
In the external-tuning track, submitters provide a workload-agnostic hyperparameter search space or discrete set of hyperparameter points. During evaluation, multiple hyperparameter configurations are sampled and run for each workload, with the final score reflecting the best-performing trial on that specific task. While external tuning quantifies the peak performance achievable when computational resources are available for hyperparameter search, it does not reflect scenarios where tuning is impractical. 

In contrast, the \emph{self-tuning ruleset} requires submitters to supply a single, fixed hyperparameter configuration that is executed across all 9 benchmark workloads. This track has a runtime budget of $1.5 \times$ the budget of the external tuning track.
Submissions are free to define hyperparameters and adjust them while training based on some heuristic or procedure that can generalize to new workloads, but their score will be a function of how long they run and will include any time spent trying out different configurations.

\paragraph{Importance for practical deep learning adoption.}
Evaluating optimizers under the self-tuning ruleset is vital for the practical adoption of new optimizers. For large-scale training jobs, hyperparameter sweeps are often expensive in terms of computational budget, energy consumption, and turnaround time. Practitioners require easy-to-use optimizers that aren't finicky about particular hyperparameter settings, and the self-tuning track strongly rewards such optimizers.

\subsection{Scoring}

A submission’s AlgoPerf benchmark score is based on its individual workload scores
relative to those of other submissions, aggregated using performance profiles. A performance profile plots the fraction of \wls where a submission trains successfully (achieves the target performance) within a factor of $\tau$ of the time required by the per-\wl fastest submission, for different values of $1 \!\leq\! \tau \!\leq\! \tau_{\text{max}}$.
For instance, the height of submission $s$'s performance profile at $\tau\!=\!1.5$ is the fraction of \wls where $s$ trains successfully and within $1.5\!\times$ the wall-clock time the best competitor requires on that \wl.
The final scalar \emph{benchmark scores} are the normalized area under the performance profiles, with $1.0$ corresponding to a submission that trained the fastest on all \wls.

\section{\algoperf leaderboard}
Below we present the results of the submissions to the \algoperfnew self-tuning leaderboard. We first summarize the evaluated submissions and their leaderboard rankings according to wall-clock time, and then examine how the rankings shift under step-based scoring and relaxed validation targets.

\subsection{Evaluated algorithms}
The self-tuning leaderboard evaluates a diverse set of first-order, momentum-augmented, and matrix-preconditioned optimization algorithms submitted by the community and working group members. Table~\ref{tab:evaluated_algorithms} summarizes all evaluated submissions, providing brief notes, citations, and direct links to the submission code repositories.

\begin{table*}[htbp]
  \centering
  \footnotesize
  \caption{Summary of evaluated algorithms on the \algoperf self-tuning leaderboard.}
  \label{tab:evaluated_algorithms}
  \setlength{\tabcolsep}{4pt}
  \begin{tabularx}{\textwidth}{@{}>{\raggedright\arraybackslash}p{2.9cm} >{\raggedright\arraybackslash}X >{\raggedright\arraybackslash}p{2.7cm} >{\raggedright\arraybackslash}p{3.9cm}@{}}
    \toprule
    \textbf{Algorithm} & \textbf{Brief Description} & \textbf{Citation} & \textbf{Submission Code} \\
    \midrule
    \sfadamwii & Couples momentum interpolation with dynamic iterate averaging; eliminates LR schedules and maintains batch norm statistics. This version makes use of the batch-norm fixes in \algoperfnew and is as a result more faithful to the published version of Schedule-Free. & \citep{defazio2024roadscheduled} & \href{https://github.com/mlcommons/submissions_algorithms/tree/main/submissions/self_tuning/schedule_free_adamw_v2}{\texttt{schedule\_free\_adamw\_v2}} \\
    \ademamixadamw & Blends fast and slow gradient EMAs to track short-term fluctuations while retaining long-term directional memory. This submission sets a long warmup for $\alpha$ and $\beta_3$ making it effectively equivalent to AdamW. & \citep{pagliardini_the_adem_opti24} & \href{https://github.com/mlcommons/submissions_algorithms/tree/main/submissions/self_tuning/ademamix}{\texttt{ademamix}} \\
    \sfadamw & Legacy Schedule-Free AdamW implementation submitted to the \algoperf competition. This version does not contain batch-norm fixes. & \citep{defazio2024roadscheduled} & \href{https://github.com/mlcommons/submissions_algorithms/tree/main/submissions/self_tuning/schedule_free_adamw}{\texttt{schedule\_free\_adamw}} \\
    \nadamwbase & Nesterov-accelerated adaptive gradient optimizer with decoupled weight decay. This version served as the prize qualification baseline for the AlgoPerf v0.5 competition. & \citep{dozat2016incorporating,loshchilov2018decoupled} & \href{https://github.com/mlcommons/submissions_algorithms/tree/main/submissions/self_tuning/nadamw_baselinev05}{\texttt{nadamw\_baselinev05}} \\
    \sfadamwjaxii & JAX implementation of Schedule-Free AdamW v2 with batch-normalization fix. & \citep{defazio2024roadscheduled} & \href{https://github.com/mlcommons/submissions_algorithms/tree/main/submissions/self_tuning/schedule_free_adamw_jax_v2}{\texttt{schedule\_free\_adamw\_jax\_v2}} \\
    \muonpt & Matrix-preconditioned optimizer applying quintic Newton-Schulz iterations on 2D or higher-dimensional non-embedding tensors with AdamW for 1D and embedding parameters. & \citep{jordan2024muon} & \href{https://github.com/mlcommons/submissions_algorithms/tree/main/submissions/self_tuning/muon_torch}{\texttt{muon\_torch}} \\
    \sfadamwjax & JAX implementation of the legacy Schedule-Free AdamW submission to the \algoperf v0.5 competition without batch-norm fixes. & \citep{defazio2024roadscheduled} & \href{https://github.com/mlcommons/submissions_algorithms/tree/main/submissions/self_tuning/schedule_free_adamw_jax}{\texttt{schedule\_free\_adamw\_jax}} \\
    \nadamw & Nesterov-accelerated adaptive gradient optimizer with decoupled weight decay. & \citep{dozat2016incorporating,loshchilov2018decoupled} & \href{https://github.com/mlcommons/submissions_algorithms/tree/main/submissions/self_tuning/nadamw}{\texttt{nadamw}} \\
    \cautiousnadamw & Directional consistency optimizer that executes parameter updates only when current gradient and momentum share the same sign. & \citep{liang2024cautious} & \href{https://github.com/mlcommons/submissions_algorithms/tree/main/submissions/self_tuning/cautious_nadamw}{\texttt{cautious\_nadamw}} \\
    \muonjax & JAX implementation of the Newton-Schulz matrix preconditioning algorithm. & \citep{jordan2024muon} & \href{https://github.com/mlcommons/submissions_algorithms/tree/main/submissions/self_tuning/muon}{\texttt{muon}} \\
    \lion & Memory-efficient sign-based momentum optimizer with uniform coordinate update magnitudes. & \citep{chen2023symbolic} & \href{https://github.com/mlcommons/submissions_algorithms/tree/main/submissions/self_tuning/lion}{\texttt{lion}} \\
    \ademamix & Blends fast and slow gradient EMAs to track short-term fluctuations while retaining long-term directional memory. & \citep{pagliardini_the_adem_opti24} & \href{https://github.com/mlcommons/submissions_algorithms/tree/main/submissions/self_tuning/ademamix_golden}{\texttt{ademamix\_golden}} \\
    \dilocosw & Hierarchical two-loop optimization with inner AdamW steps and outer Nesterov momentum synchronization. & \citep{douillard2023diloco} & \href{https://github.com/mlcommons/submissions_algorithms/tree/main/submissions/self_tuning/single_worker_diloco}{\texttt{single\_worker\_diloco}} \\
    \bottomrule
  \end{tabularx}
\end{table*}

\clearpage
\subsection{\algoperf scores}

We present the main \algoperf leaderboard. Submissions are ranked in \cref{tab:scores},\footnote{The second-place submission, \ademamixadamw, uses a $500{,}000$-step slow-EMA warmup that renders it effectively equivalent to AdamW on the benchmark workloads (see \cref{sec:ademamix_analysis}).} and their performance profiles are shown in \cref{fig:performance_profiles}.
Per-workload times to target are reported in \cref{tab:time_to_target} in \cref{app:step_execution_times}. The \sfadamwii submission ranked first on the leaderboard followed by \ademamixadamw. As noted in Table \ref{tab:evaluated_algorithms} the \ademamixadamw submission sets its hyperparameter such that it is essentially equivalent to AdamW, making this submission just a very well-tuned AdamW submission. The performance profiles show that the top-ranking submission \sfadamwjaxii achieved fast ($<1.25 \tau$) time to target on 5 of the 9 workloads and was reached the target on 6 workloads total. We also note that the top-3 submissions were PyTorch submissions and that the equivalent JAX version of the top-scoring submission \sfadamwjaxii reached the target on a different set of workloads. We believe that this is due to workload runtime differences between PyTorch and JAX, details of which are discussed in \cref{sec:systems_analysis}.

\leaderboardprofilefigure
    {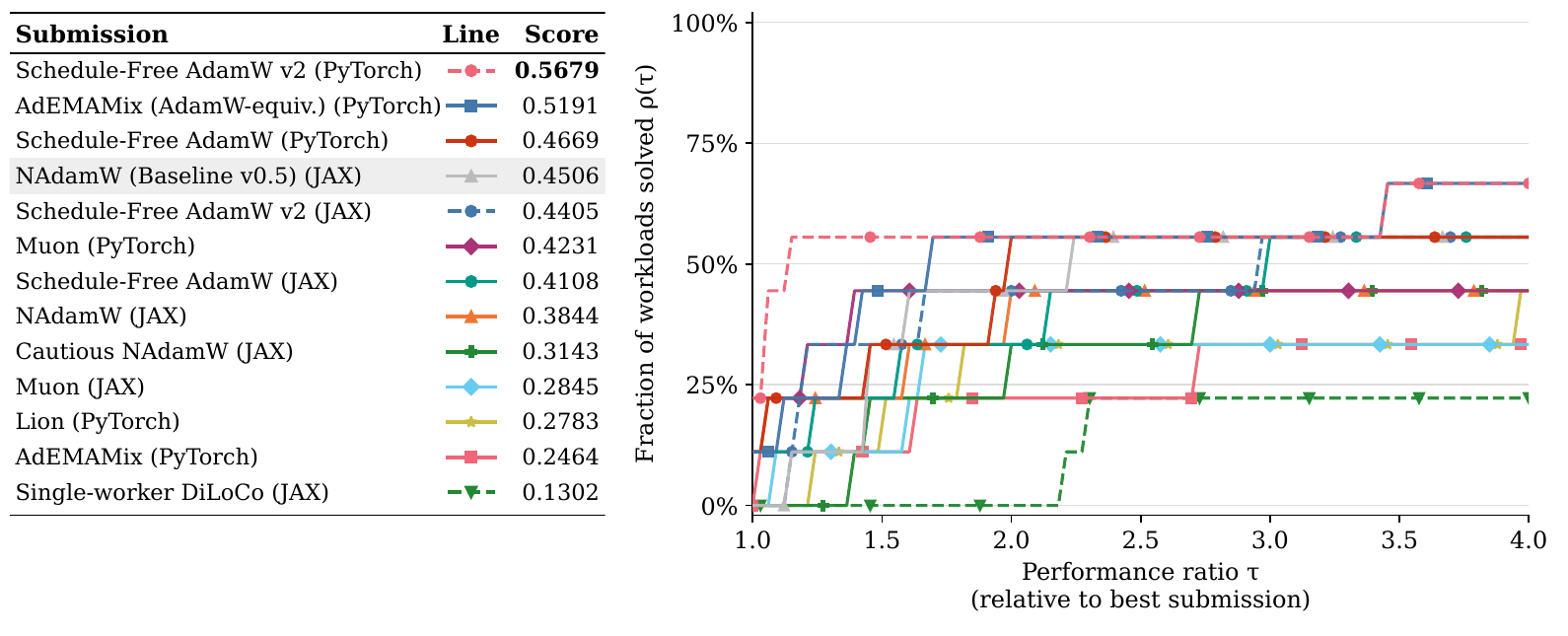}
    {AlgoPerf self-tuning leaderboard and performance profiles. The leaderboard ranks the submissions according to their \algoperf scores, which correspond to the areas under the curves of the performance profiles. }
    {fig:leaderboard_profiles}
    {tab:scores}
    {fig:performance_profiles}

\subsection{\algoperf wall-clock vs step efficiency}
A central tenet of the \algoperf benchmark design is evaluating optimization algorithms on a \emph{wall-clock} leaderboard rather than a \emph{step-based} (or sample-efficiency) leaderboard. The former is representative of the real-world performance of a given algorithm, but the latter gives insight into an optimizer's per-step optimization efficiency (ignoring the computational cost of each optimizer step). We analyze how the leaderboard changes based on pure step-efficiency, ignoring the computational cost of each step.
Concretely, we re-score all submissions by replacing wall-clock time-to-target with the median number of optimizer steps to target, holding all else equal. We note that submissions that did not reach the target within wall-clock time budget, still get an infinite score on those instances, making this a limitation in this analysis. In an ideal setup we would have submission data up to some step budget as well. 

\begin{figure}[htbp]
    \centering
    \includegraphics[width=\textwidth]{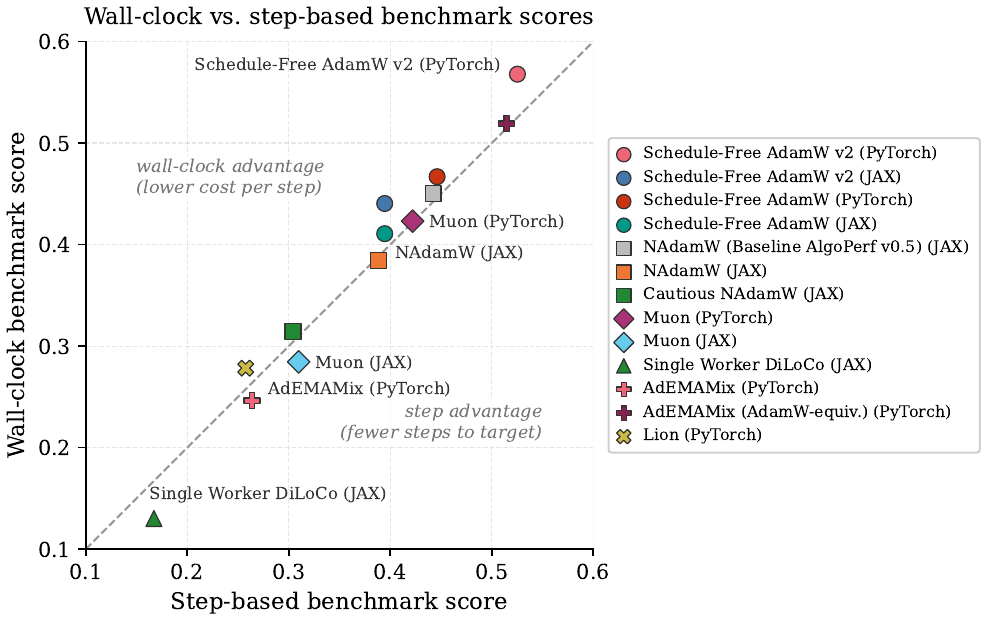}
    \caption{Wall-clock vs.\ step-based benchmark score per submission.}
    \label{fig:wallclock_vs_steps}
\end{figure}

We compare the two scores in \cref{fig:wallclock_vs_steps}, revealing which algorithms perform better due to requiring fewer steps to target, at the potential cost of more expensive computation per step. Algorithms that are step-efficient but expensive in wall-clock, like DiLoCo and \muonjax{}, receive slightly better scores under the step-based scoring compared to time-based scoring. This means that while their step times might be computationally more expensive, they take better steps. For Muon specifically, we see that more optimized implementations like \muonpt can result in better performance on the wall-clock based leaderboards. A more comprehensive analysis of Muon specifically is in \cref{sec:muon_analysis}. 

\cref{tab:leaderboard_steps} in \cref{app:step_based_leaderboard} reports both leaderboards side by side (with per-workload step counts in \cref{tab:steps_to_target}), and \cref{fig:step_count_profiles} shows the step-based performance profiles.

\subsection{Relaxed \algoperf leaderboards}

If a submission fails to achieve the validation target on a workload it will not receive a finite time-to-target score on that workload and as a result does not contribute anything to the final \algoperf score. That procedure, however, penalizes algorithms that \emph{almost} reached the target, but did not succeed. In reality, an algorithm may still be useful if it nearly reaches the target, depending on the workload. So we can evaluate on ``relaxed'' targets by making them some percentage easier than the default targets. Concretely, for a relaxation fraction $r$, each workload's target $\tau$ becomes $(1+r),\tau$ for metrics that are minimized (loss, WER, perplexity) and $(1-r),\tau$ for metrics that are maximized (accuracy, SSIM, mAP, BLEU), with the same $r$ applied to every workload.
One limitation of this analysis is that because the targets lie at different distances from where each metric saturates, the same relative relaxation is not equally lenient across workloads; for example, with $r = 10\%$ the baseline NAdamW reaches the relaxed \criteo target in about 5\% of the time it needs for the default target, compared with about 75\% on \finewebedu.

\begin{figure}[htbp]
    \centering
    \includegraphics[width=\textwidth]{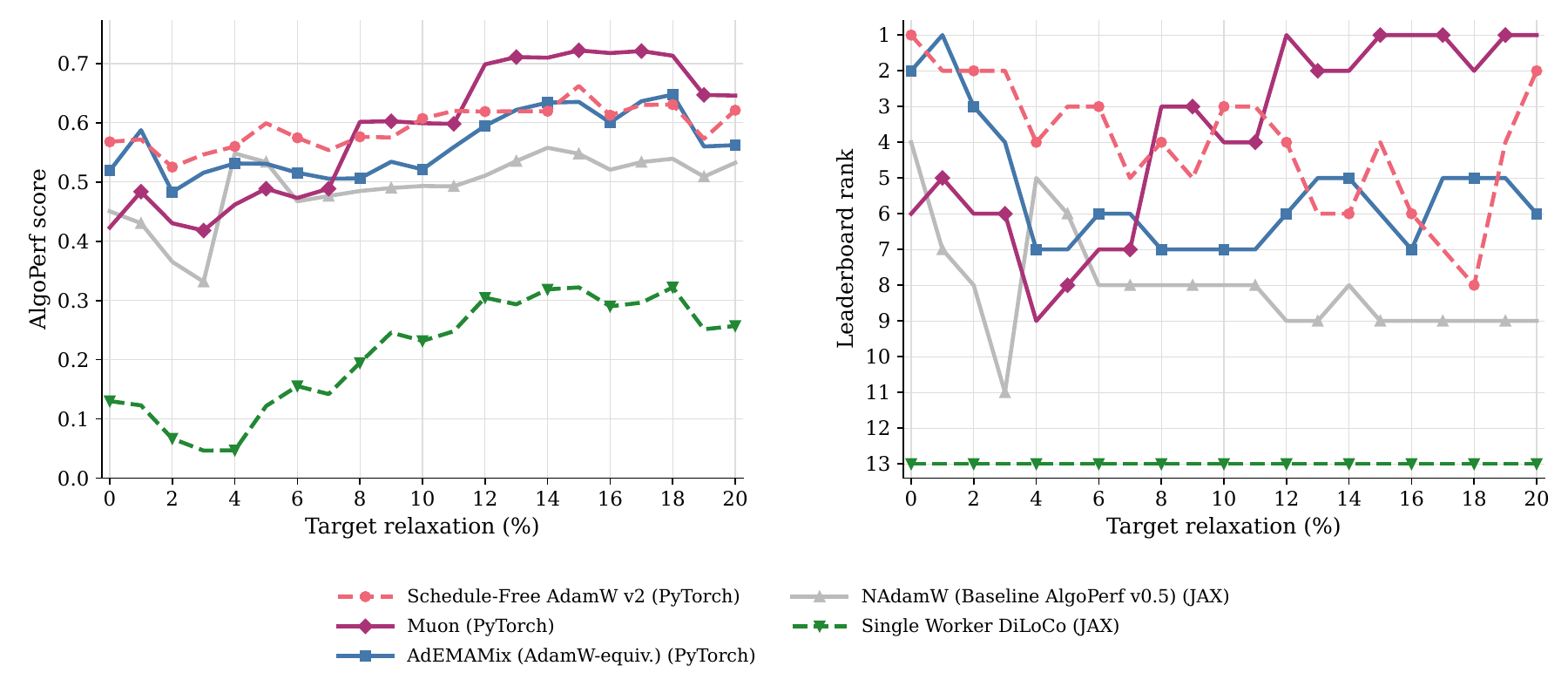}
    \caption{\algoperf score (left) and leaderboard rank (right) as targets are relaxed.}
    \label{fig:target_relaxation_sweep}
\end{figure}

We select a few interesting algorithms that were close to reaching the validation target on many workloads, and plot their leaderboard score and rank with respect to a relaxation percentage applied to the workload targets (\cref{fig:target_relaxation_sweep}).
In particular, \muonpt climbs the leaderboard very quickly as we relax the targets climbing from 6th place to 1st for a 12\% relaxation. \sfadamwii performs consistently well across all levels of relaxation, while \dilocosw is provided as a ``floor'' score.

Overall, we also observe that the \algoperf scores are more stable with respect to the targets, compared to the leaderboard rankings. Full relaxed leaderboard profiles and per-workload times to target at a 10\% target relaxation are provided in \cref{app:relaxed_leaderboard} (\cref{tab:scores_relaxed} and \cref{tab:time_to_target_relaxed}).

\section {Tuning}
\label{sec:tuning}
Because the self-tuning ruleset requires a single hyperparameter configuration to succeed across all nine workloads, understanding hyperparameter sensitivity is essential for interpreting leaderboard performance. We divide this investigation into two complementary analyses: \cref{sec:saliency_analysis} identifies \emph{which individual hyperparameters are the most sensitive} and whether their top-performing intervals overlap across workloads, while \cref{sec:tuning_overhead} quantifies the \emph{overall tuning overhead and cross-workload retention} when moving from per-workload oracle tuning to a single universal configuration. We evaluate these questions using halton sweeps of $N = 100$ configurations across all nine \algoperf workloads for Single-Worker DiLoCo, NAdamW, Muon, and AdEMAMix (see Appendix~\ref{sec:appendix_sweep_details} for search space details).

\subsection{Hyperparameter saliency analysis}
\label{sec:saliency_analysis}
An ideal, easy-to-tune training algorithm requires minimal adjustment across different tasks. In practice, however, most training algorithms rely on critical hyperparameters that are sensitive to the specific workload. To quantify this sensitivity, we measure hyperparameter importance using the 1D Wasserstein distance between the top~$k\%$ configurations and the overall sampling distribution. Since the metric we define below depends on the search space, we selected the overall sampling distribution such that it contains good samples for all 9 \algoperf workloads.

Specifically for a given workload $w$, optimizer, and hyperparameter tuning interval $\theta \in [\theta_{\min}, \theta_{\max}]$, let $P_{\text{all}}(\theta)$ be the empirical distribution of hyperparameter values sampled across all tuning trials in the search space.
Let $P_{\text{top-}k}(\theta)$ be the conditional distribution of $\theta$ restricted to trials achieving top $k\%$ performance (e.g., top $10\%$ of final evaluation metrics). We then define the normalized Importance Index $\mathcal{I}(\theta)$ as the 1-Wasserstein distance between the top-performing trials and the prior proposal distribution, normalized by the total search span:
\begin{equation}
  \mathcal{I}(\theta) = \frac{\mathcal{W}_1\left(P_{\text{top-}k}(\theta),\, P_{\text{all}}(\theta)\right)}{\theta_{\max} - \theta_{\min}}
  \label{eq:importance_index}
\end{equation}
For parameters sampled over log scales (such as learning rate and weight decay), distributions and distances are evaluated in $\log_{10}$ space. So intuitively, $\mathcal{I}(\theta)$ measures how much distributional shift is required for top-k performance. We include visualizations of these top-K distributions in \cref{sec:topk_distributions}. Low $\mathcal{I}$ indicates that the algorithm is not very sensitive to the variations in the hyperparameter, while a high $\mathcal{I}$ indicates that precise tuning is necessary for a given hyperparameter within the tuning interval.

\begin{figure}[t]
  \centering
  \includegraphics[width=\linewidth]{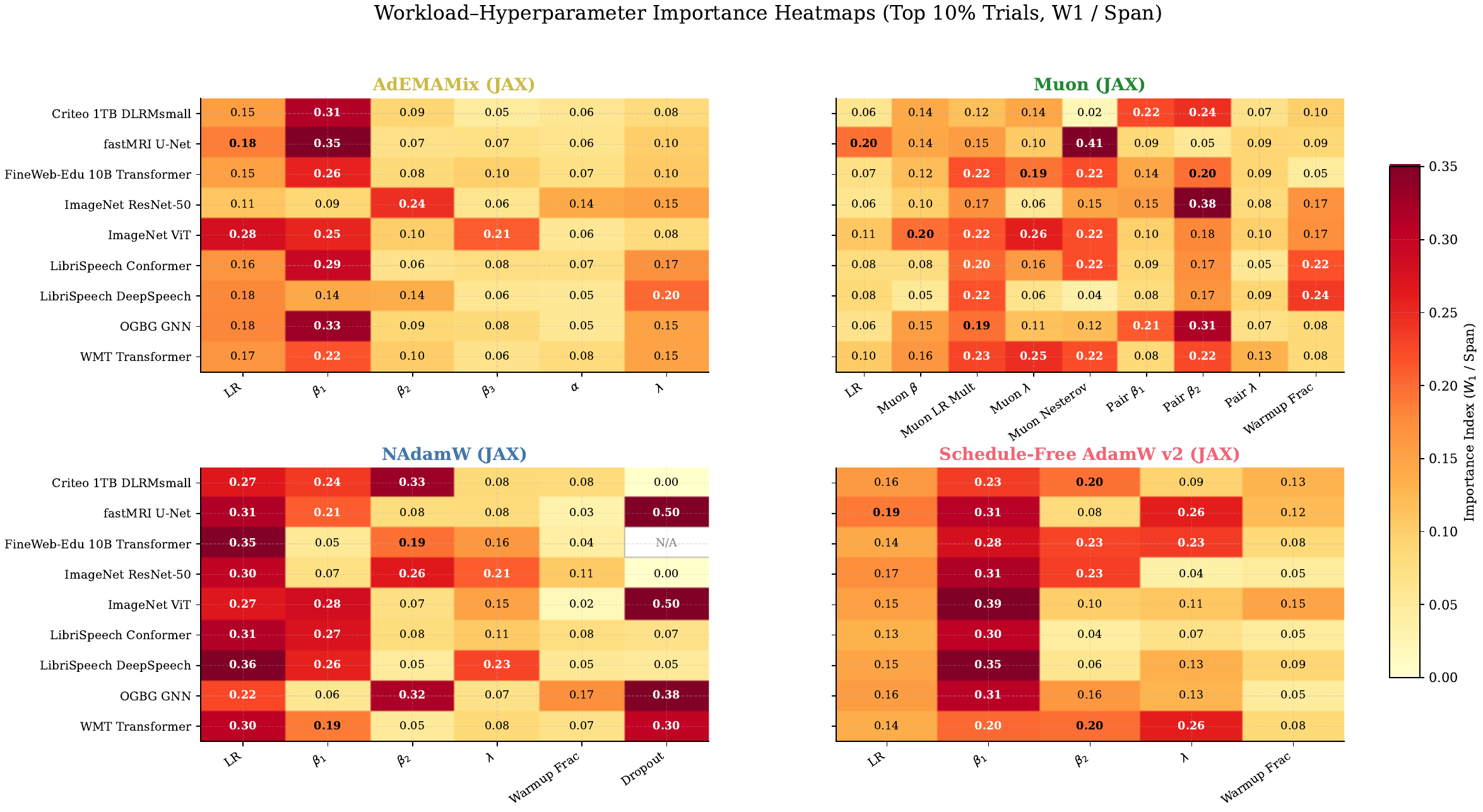}
  \caption{\textbf{Workload--Hyperparameter Importance Heatmaps Across Optimizers.}
  Normalized 1D Wasserstein importance indices ($\mathcal{W}_1 / \text{Span}$, Equation~\ref{eq:importance_index}) across the AlgoPerf benchmark suite, evaluated at the top $10\%$ best-performing trial threshold ($k=0.10$). Higher values (darker red) denote hyperparameters that have to collapse into a narrow band to achieve top-tier performance, while lower values reflect broad robustness across the search space. Each subplot isolates the active hyperparameters specific to that optimizer.}
  \label{fig:hparam_importance_heatmaps}
\end{figure}

As shown in Figure~\ref{fig:hparam_importance_heatmaps}, inspecting the importance indices across both axes yields several key observations:
\begin{itemize}
  \item \textbf{Universality of Learning Rate Sensitivity:} The base learning rate ($\eta$) consistently ranks among the top 2 most important hyperparameters across workloads and algorithms.
  \item \textbf{Momentum sensitivity.} Most of the algorithms also show particular sensitivity to the momentum $\beta_1$ or $\beta$ (for Muon).
\end{itemize}

\begin{figure}[htbp]
    \centering
    \includegraphics[width=\linewidth]        {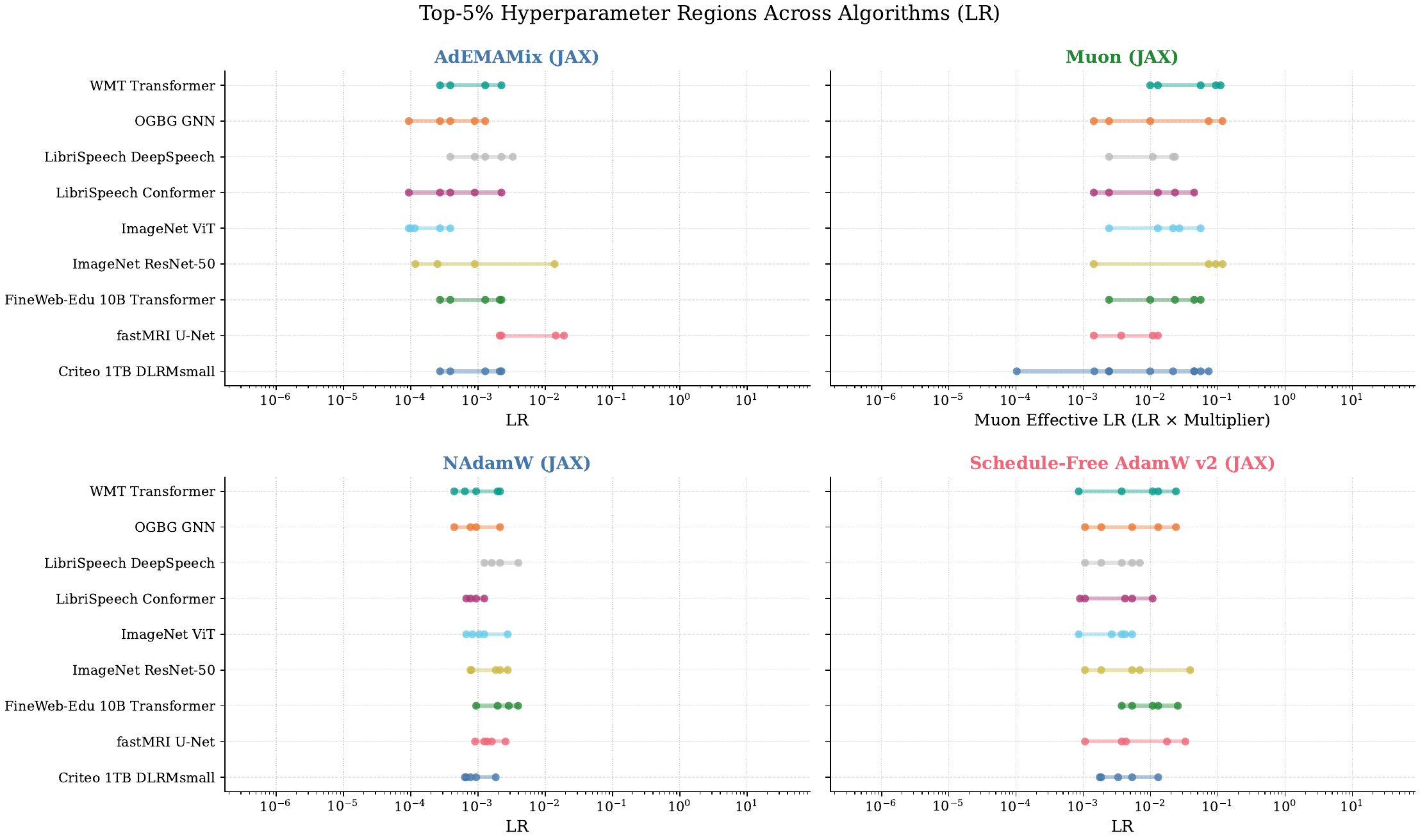}
    \caption{Spans of learning rate across the top 5\% configurations. Disjoint regions indicate fundamental tension for universal tuning. Note that for Muon, the LR indicates the learning rate applied to the paired optimizer and serves as a basis for the Muon LR multiplier.}
    \label{fig:topk_lr}
\end{figure}

We next investigate the difficulty of identifying hyperparameter configurations along the most sensitive axes that generalize effectively across workloads. In the context of the self-tuning track, this represents the core challenge, since most submissions rely on empirical default values for their hyperparameters. The top 5\% learning rate spans visualized in \cref{fig:topk_lr} demonstrate that this alignment can be quite challenging in practice. For NadamW the top-5\% \librispeech \conformer and \librispeech \deepspeech points are nearly disjoint. For AdEMAMix, this discrepancy is even more pronounced: the top configurations for \fastmri \unet are nearly disjoint from three workloads and entirely disjoint from two others.

\FloatBarrier
\subsection{Hyperparameter tuning overhead and retention analysis}
\label{sec:tuning_analysis}
\label{sec:tuning_overhead}

Evaluating optimizers solely on their submitted configurations penalizes submissions that can't specify a single configuration that does well across workloads, but doesn't reveal all the information we would want about the costs of hyperparameter search. In practice, two questions arise: first, how easily an optimizer tunes on an individual workload, and second, how much performance is lost when deploying a single configuration across workloads without per-workload tuning. 

\paragraph{Continuous validation progress and time-based scoring.}
To measure wall-clock progress, let $T_{\max, w}$ denote the runtime budget on workload $w$ (\cref{tab:workloads}) and let $t_{\text{target}}$ denote the time to reach target $y_{\text{target}}$: either the observed crossing time $t_{\text{hit}} \le T_{\max, w}$, or, for non-diverging runs that miss the target in-sample, the projected crossing time $\hat{t}_{\text{target}}$ obtained by fitting the power law $y(t) = y_\infty \pm A\,t^{-\gamma}$ ($A, \gamma > 0$) to the running-best validation trajectory as a function of cumulative runtime $t$. Capping extrapolation at $2\,T_{\max, w}$, we define the workload speedup score as $S_w(c) = T_{\max, w} / t_{\text{target}}$ if $t_{\text{target}} \le 2\,T_{\max, w}$ and $S_w(c) = 0$ otherwise, where $c$ corresponds to a specific hyperparameter point configuration. The overall configuration score is then defined as the mean across all $W = 9$ workloads, $\bar{S}(c) = \frac{1}{W} \sum_{w=1}^W S_w(c)$.

\paragraph{Question 1: Tuning difficulty on an arbitrary workload.}
We evaluate tuning difficulty using the distribution of time-based speedup scores across each search space: median speedup ($p_{50}$) and failure rate ($P(\bar{S} \le 0.20)$).

\begin{figure}[htbp]
    \centering
    \includegraphics[width=0.98\textwidth]{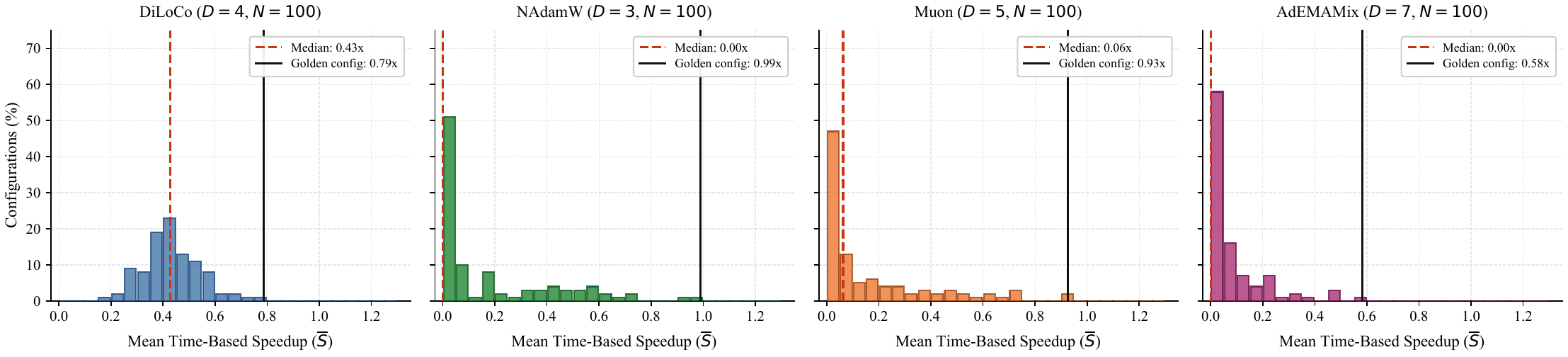}
    \caption{Distribution of nine-workload mean time-based speedup scores ($\bar{S}$) across hyperparameter sweeps for DiLoCo, NAdamW, Muon, and AdEMAMix ($N=100$ per optimizer) under a $2\times$ extrapolation horizon on the running-best validation trajectory. Vertical dashed red lines indicate medians; solid black lines mark the single golden configuration.}
    \label{fig:sensitivity_histograms}
\end{figure}

\begin{table}[htbp]
\centering
\caption{Hyperparameter sensitivity metrics across $N=100$ configurations under extrapolated time-to-target scoring ($S_w(c) = T_{\max,w} / \hat{t}_{\text{target}}$, capped at $\hat{t}_{\text{target}} \le 2\,T_{\max,w}$). Failure rate denotes $\bar{S}(c) \le 0.20$.}
\label{tab:tuning_difficulty}
\resizebox{\linewidth}{!}{%
\begin{tabular}{lrrrr}
\toprule
Tuning Metric & DiLoCo ($D=4$) & NAdamW ($D=3$) & Muon ($D=5$) & AdEMAMix ($D=7$) \\
\midrule
Golden config score ($\bar{S}(c^*)$) & 0.786 & \textbf{0.989} & 0.927 & 0.583 \\
Oracle score ($\bar{S}^*$) & 1.279 & 1.509 & \textbf{1.973} & 1.403 \\
Retention Rate ($\mathcal{R}$) & 61.4\% & \textbf{65.5\%} & 47.0\% & 41.5\% \\
Median Speedup ($p_{50}$) & \textbf{0.427} & 0.000 & 0.063 & 0.000 \\
Failure Rate ($P(\bar{S} \le 0.20)$) & \textbf{1.0\%} & 70.0\% & 71.0\% & 85.0\% \\
\bottomrule
\end{tabular}%
}
\end{table}

As shown in \cref{fig:sensitivity_histograms} and \cref{tab:tuning_difficulty}, DiLoCo remains the most robust under time-based evaluation: it incurs the lowest failure rate ($1.0\%$) and achieves the highest median speedup ($0.427\text{x}$, compared to $0.063\text{x}$ for Muon and $0\text{x}$ for NAdamW and AdEMAMix). The number of hyperparameters to tune acts as a confounding factor, yet there is no escaping it in practice: within a finite budget, algorithms with more hyperparameters naturally suffer more than algorithms with fewer hyperparameters.

\paragraph{Question 2: Relative retention.}
Question 1 quantifies how hard it is to tune an algorithm on an arbitrary workload. Question 2 asks: how much do we lose by tuning an algorithm to have a single set of hyperparameters across multiple, diverse workloads versus using per-workload optimal values? We call this retention, defined as the ratio of the golden configuration score ($\bar{S}(c^*) = \max_c \frac{1}{W} \sum_w S_w(c)$) to the per-workload oracle score ($\bar{S}^* = \frac{1}{W} \sum_w \max_c S_w(c)$), i.e., $\mathcal{R} = \bar{S}(c^*) / \bar{S}^*$.

\begin{figure}[htbp]
    \centering
    \includegraphics[width=0.98\textwidth]{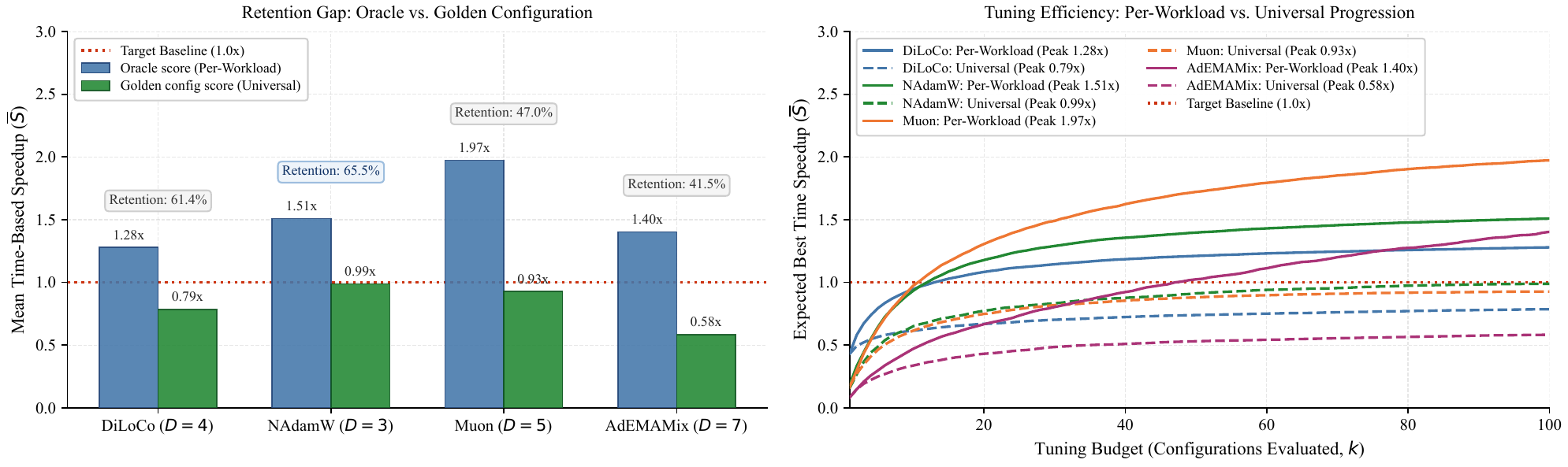}
    \caption{Retention gap and tuning budget progression across optimizers under a $2\times$ extrapolation horizon on the running-best validation trajectory. \textbf{Left:} Golden config score ($\bar{S}(c^*)$, single configuration across all 9 workloads) versus Oracle score ($\bar{S}^*$, per-workload tuning) with retention percentage $\mathcal{R} = \bar{S}(c^*) / \bar{S}^*$. \textbf{Right:} Expected best speedup $\bar{S}$ as a function of tuning budget $k \in [1, 100]$ under universal search (dashed) versus per-workload search (solid); see below.}
    \label{fig:universality_and_efficiency}
\end{figure}

As shown in \cref{fig:universality_and_efficiency} (left), while Muon achieves the highest per-workload oracle ceiling ($\bar{S}^* = 1.973$), NAdamW ($\mathcal{R} = 65.5\%$) and DiLoCo ($61.4\%$) retain the largest share of their oracle performance under a single universal configuration ($\bar{S}(c^*) = 0.989$ and $0.786$), ahead of Muon ($47.0\%$, $\bar{S}(c^*) = 0.927$) and AdEMAMix ($41.5\%$, $\bar{S}(c^*) = 0.583$). In \cref{fig:universality_and_efficiency} (right), we estimate the expected best score at budget $k$ by randomly subsampling $k$ of the $N = 100$ configurations without replacement (averaged over $500$ trials) and taking the best configuration either universally across all workloads or independently per workload. DiLoCo achieves the highest expected score at minimal budgets ($k \le 5$) due to its $1.0\%$ failure rate, whereas NAdamW and Muon overtake it as the budget grows by drawing from their upper tails, and AdEMAMix requires substantially larger budgets to approach its golden configuration.

\section{Submission Implementation Case Studies}
\label{sec:algorithm_analysis}
Aggregate benchmark rankings compress each of the 13 submissions into a single scalar score, and can obscure the low-level implementation choices that dictate actual runtime behavior. In this section, we examine three case studies that illustrate how optimizer implementation and design details dictated benchmark success or failure:
\begin{enumerate}
    \item \textbf{Schedule-Free AdamW}: How decoupling running-statistic updates from parameter updates eliminated severe Batch Normalization covariate shifts on \resnet \imagenet (\cref{sec:schedule_free_analysis}).
    \item \textbf{Muon}: How distributed tensor sharding and asynchronous communication overlap mitigated the steep per-step orthogonalization latency on parameter-dense workloads (\cref{sec:muon_analysis}).
    \item \textbf{AdEMAMix}: How long warmups for momentum hyperparameters can lead to accidental collapse into AdamW. (\cref{sec:ademamix_analysis}).
\end{enumerate}

\subsection{Schedule-Free AdamW: Batch Normalization Calibration}
\label{sec:schedule_free_analysis}

Schedule-Free optimization \citep{defazio2024roadscheduled} eliminates manual learning rate decay schedules by interpolating between Polyak-Ruppert and primal iterate averaging. The algorithm maintains three parameter sequences: a base sequence $z_t$, an evaluation sequence $x_t$ (a running average of $z_t$), and an intermediate sequence $y_t = (1 - \beta) x_t + \beta z_t$ where gradients are evaluated \citep{defazio2024roadscheduled}. This formulation removes the requirement of pre-specifying the total step budget $T$ for a learning rate decay schedule.

\paragraph{Empirical performance across workloads.}
Schedule-Free AdamW demonstrated the broadest out-of-the-box generalization across the suite. On the self-tuning leaderboard, \sfadamwii placed first, reaching the validation target on six workloads. This broad applicability confirms that schedule-free iterate averaging can succeed across heterogeneous architectures without task-specific schedule tuning. While \sfadamwii did not reach the target on \resnet \imagenet, this submission required diagnosing and fixing an batch normalization bug that was crucial for stabilizing training on the \resnet workload.

\paragraph{Resolving covariate shift in Batch Normalization buffers.}
In standard optimization, the parameters used for forward propagation during training are identical to those used during evaluation. In the original \algoperf submission (\sfadamw), Batch Normalization (BN) running statistics $(\hat{\mu}, \hat{\sigma}^2)$ were accumulated during the forward-backward pass at the extrapolated weights $y_t$. However, model evaluation is performed on the averaged weights $x_t$. Because $y_t \neq x_t$, evaluating $x_t$ with running statistics calibrated for $y_t$ induces a covariate shift across normalization layers that severely degrades validation accuracy on BN-dependent tasks.

\begin{figure}[htbp]
    \centering
    \includegraphics[width=0.82\textwidth]{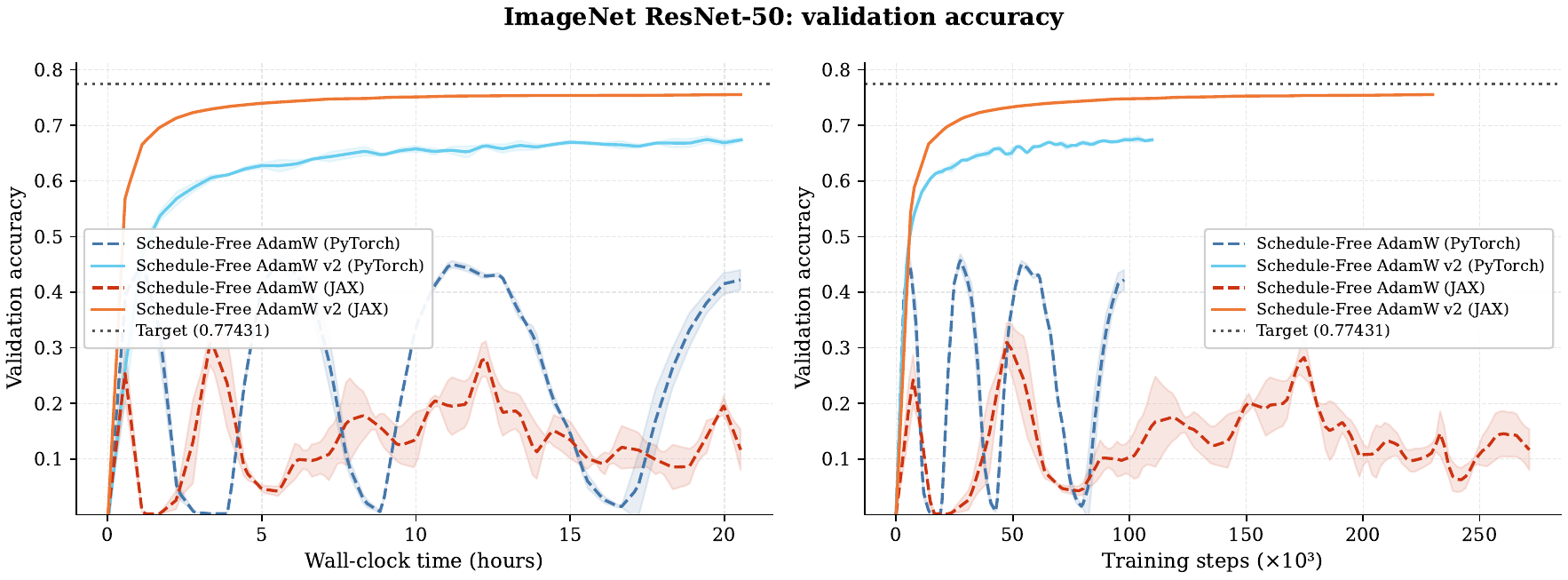}
    \caption{Training dynamics of Schedule-Free AdamW on the ImageNet \resnet workload. Calibrating Batch Normalization running statistics at $x_t$ rather than $y_t$ in \sfadamwii resolves the severe covariate shift that stalled validation progress in earlier iterations.}
    \label{fig:sfadamw_resnet}
\end{figure}

To eliminate this mismatch, \sfadamwii decouples gradient flow from buffer tracking:
\begin{enumerate}
    \item The training forward-backward pass executes at $y_t$ using strictly in-batch statistics, with running buffer updates disabled to ensure clean gradient evaluation.
    \item Running statistics are independently calibrated to $x_t$ by executing an additional forward pass (with gradient tracking disabled) at the evaluation weights $x_t$ every third training step.
\end{enumerate}
In addition, \sfadamwii applies weight decay directly to the extrapolated point $y_t$ (rather than the base sequence $z_t$) and replaces the $r = 0.5$ factorial power weighting sequence used in v1 with a straightforward learning rate-weighted average. As illustrated in \cref{fig:sfadamw_resnet}, calibrating the BN buffers directly at $x_t$ stabilizes validation progress on \resnet and \deepspeech, transforming ImageNet ResNet-50 from a failure mode into a stable run.

\FloatBarrier
\Needspace{12\baselineskip}
\subsection{Muon: Distributed Sharding and Computational Overhead}
\label{sec:muon_analysis}

Muon \citep{jordan2024muon} replaces coordinate-wise Euclidean updates on matrix parameters with spectral orthogonalization of the momentum matrix $\bm{M}_t \in \mathbb{R}^{m \times n}$:
\begin{equation}
    \bm{O}_t = \bm{M}_t (\bm{M}_t^T \bm{M}_t)^{-1/2},
    \label{eq:muon_polar}
\end{equation}
which normalizes spectral energy across singular modes. To avoid the prohibitive $O(\min(m, n)^3)$ cost of exact singular value decompositions at every step, practical implementations approximate the polar factor in \cref{eq:muon_polar} using a quintic ($5$-th order) Newton-Schulz polynomial:
\begin{equation}
    \bm{X}_{k+1} = a\,\bm{X}_k + \bm{X}_k \left( b\,\bm{X}_k^T \bm{X}_k + c\,(\bm{X}_k^T \bm{X}_k)^2 \right)
    \label{eq:newton_schulz}
\end{equation} 
Parameters are partitioned according to their structure: all embedding tables and 1D parameters are updated via standard AdamW/NAdamW, while all remaining tensors with $\text{ndim} \ge 2$ are routed to Muon. For tensors with three or more dimensions (such as 4D convolutional kernels in \resnet), the trailing dimensions are flattened into a single dimension ($m \times \prod_{i=1}^d n_i$) prior to Newton-Schulz orthogonalization and restored to their original shapes before the weight update.

\paragraph{Distributed sharding vs.\ replicated computation.}
In the reference implementation of \citet{jordan2024muon} (vanilla Muon) as well as \muonjax, the full Newton-Schulz orthogonalization is replicated identically across all GPUs. Because each Newton-Schulz iteration requires $2m^2n + m^3$ FLOPs (for $m \le n$), this redundant linear algebra introduces substantial wall-clock latency on architectures dominated by large weight matrices.

To eliminate this bottleneck, the submitted \muonpt implementation introduces work-balanced parameter sharding:
\begin{enumerate}
    \item \textbf{Greedy Complexity Partitioning:} Matrices are sorted in descending order of their Newton-Schulz FLOP complexity ($2m^2n + m^3$) and assigned to GPUs in round-robin order, balancing orthogonalization workloads across devices.
    \item \textbf{Asynchronous Collective Communication:} Gradients are averaged and distributed to their assigned GPU via an internal \texttt{reduce-scatter}, while the resulting parameter updates are synchronized across ranks using an asynchronous \texttt{all-gather}.
    \item \textbf{Reduced-Precision Compilation:} Newton-Schulz iterations are cast to \texttt{bfloat16} and fused via \texttt{torch.compile}, in contrast to \muonjax which evaluates in \texttt{float32}.
\end{enumerate}

\begin{figure}[htbp]
    \centering
    \includegraphics[width=\textwidth]{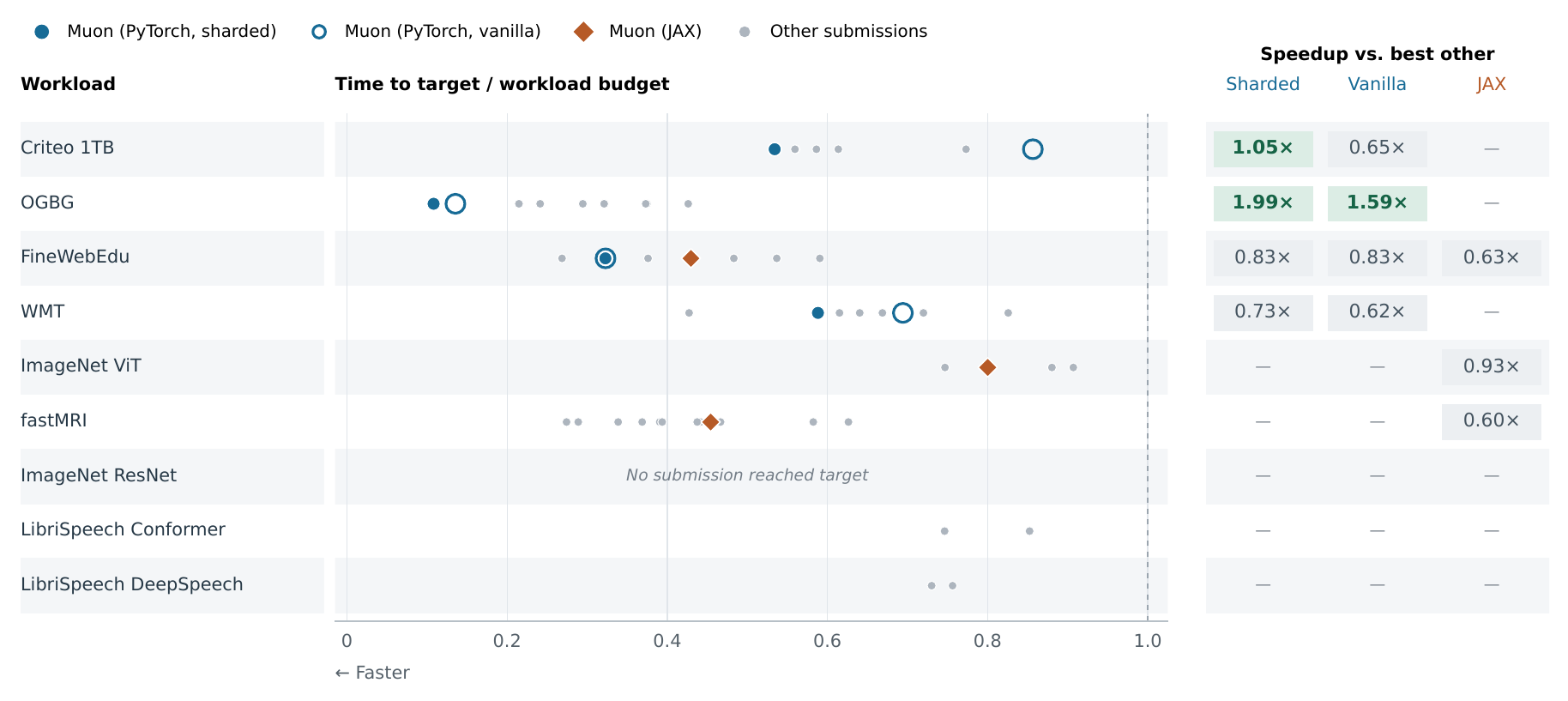}
    \caption{Empirical speedup across workloads for Muon implementations, contrasting the communication-sharded \muonpt against the replicated baseline.}
    \label{fig:muon_workloads}
\end{figure}

\paragraph{Workload sensitivity to orthogonalization overhead.}
As shown in the ablation of \cref{tab:muon_waterfall} and the step-time measurements in \cref{tab:step_time_comparison}, the speedup achieved by sharding depends directly on the model's architectural profile:
\begin{itemize}
    \item \textbf{Matrix-dominated workloads (\criteo, \wmt):} In recommendation and machine translation models, parameters are concentrated in massive linear layers. Here, orthogonalization consumes a significant fraction of step time in the replicated baseline, and sharding yields dramatic wall-clock reductions that enable \muonpt to reach the validation targets within budget (\cref{fig:muon_workloads}).
    \item \textbf{Compute-dominated workloads (\finewebedu, \ogbg):} In large transformer and graph neural network models, forward and backward propagation dominate total step execution time, rendering the Newton-Schulz computation relatively minor and making the gains from sharding negligible.
\end{itemize}
\cref{tab:muon_waterfall} demonstrates that both hyperparameter and framework runtime overhead account for performance discrepancy between \muonpt and \muonjax. Systems optimizations like sharding lowered per-step latency enough to allow good hyperparameter configurations to train to completion within the wall-clock budgets.

\begin{table}[htbp]
  \centering
  \caption{Hyperparameter ablations on Muon submissions. In this table we compare the hypothetical scores for \muonjax and \muonpt submissions ablations in the overall submission pool. For the \muonpt submissions we evaluate both the vanilla and sharded version on a combination of the hyperparameters found for the vanilla (JAX) implementation and the sharded (Torch) implementations. Each row changes one factor (bold) relative to the row above. Rows one and four are the leaderboard entries \muonjax{} and \muonpt{}; the final row steps back down to the vanilla implementation.}
  \label{tab:muon_waterfall}
  \begin{tabular}{lllrr}
    \toprule
    Framework & Implementation & HPs & Score & $\Delta$ \\
    \midrule
    JAX & vanilla & JAX & $0.2845$ & --- \\
    \textbf{PyTorch} & vanilla & JAX & $0.2969$ & $+0.012$ \\
    PyTorch & \textbf{sharded} & JAX & $0.3137$ & $+0.017$ \\
    PyTorch & sharded & \textbf{Torch} & $0.4231$ & $+0.109$ \\
    PyTorch & \textbf{vanilla} & Torch & $0.3827$ & $-0.040$ \\
    \bottomrule
  \end{tabular}
\end{table}

\subsection{AdEMAMix: Slow-EMA Warmup Horizons and AdamW Equivalence}
\label{sec:ademamix_analysis}

AdEMAMix \citep{pagliardini_the_adem_opti24} augments AdamW's fast momentum buffer ($\beta_1$) with an auxiliary slow gradient EMA ($\beta_3$), blending the two estimators via a mixing weight $\alpha(t)$ that linearly ramps over a warmup period $T_{\text{warmup}}$.
\paragraph{The 500k-step warmup collapse.}
The second-place submission, \ademamixadamw (score $0.5191$), set a fixed warmup constant of $T_{\text{warmup}} = 500{,}000$. However, all nine \algoperf workloads operate on substantially shorter horizons, with step hints in the $8\text{k}$ to $113\text{k}$ range. Because total step counts never approached $T_{\text{warmup}}$, $\alpha(t)$ remained near zero throughout training, and $\beta_3(t) \approx \beta_1(t)$. Consequently, the slow EMA was never meaningfully engaged: the submission performed well precisely because it degenerated into a well-tuned AdamW baseline with a slightly rescaled step size, rather than demonstrating the utility of the AdEMAMix mechanism itself.
\paragraph{Budget-scaled warmup and cross-workload brittleness.}
In contrast, the canonical \ademamix submission (Rank 12, score $0.2464$) attempted to genuinely engage the slow EMA across all tasks by scaling the warmup horizon to each workload's step budget ($T_{\text{warmup}} = 0.60 \times \text{step\_hint}$), derived from the universal golden configuration in \cref{sec:tuning_overhead}. 

While this budget-aware scaling succeeded in activating the slow EMA—achieving rapid convergence on language modeling (\finewebedu, reaching the target in $0.27\times$ budget vs.\ $0.38\times$ for the baseline)—it did not do well on other workloads. This contrast reflects the core findings of our hyperparameter retention analysis (\cref{sec:tuning_overhead}): AdEMAMix exhibits relatively low cross-workload retention ($\mathcal{R} = 41.5\%$). Having 7 effective hyperparameters makes it considerably more difficult to tune compared to its 5 hyperparameter AdamW counterpart. 


\section{Conclusion}

We present the results of the updated \algoperfnew self-tuning leaderboard across 13 training algorithm submissions implemented in PyTorch and JAX. Compared to \algoperfold, we evaluate several new methods on upgraded hardware ($4\times\text{A100}$ vs.\ $8\times\text{V100}$ GPUs) with recalibrated runtime budgets and a new decoder-only language modeling workload on \finewebedu. On the new leaderboard, \sfadamwii places first and retains its crown from the first benchmark. Beyond aggregate wall-clock rankings, we also analyze three aspects of optimizer behavior across the benchmark:
\begin{enumerate}[label=\alph*.]
    \item \textbf{Step efficiency vs.\ wall-clock execution:} Comparing the wall-clock leaderboard against a step-based leaderboard and relaxed target thresholds shows that preconditioned and two-loop methods like Muon and Single-Worker DiLoCo make strong per-step progress, but require systems optimizations like parameter sharding to offset their step latency.
    \item \textbf{Hyperparameter saliency and transfer:} Using a 1D Wasserstein distance metric on top-$k$ trials, we show that learning rate and momentum dominate optimizer hyperparameter sensitivity, and that top-performing learning rate intervals are often disjoint across workloads.
    \item \textbf{Tuning overhead and retention:} By comparing per-workload oracle configurations against a single universal ``golden'' configuration across hyperparameter sweeps, we find that algorithms like Muon benefit substantially more from per-workload tuning than Single-Worker DiLoCo or NAdamW, which retain a larger share of their specialized performance under a single default.
\end{enumerate}
Our results show that there is a lot of room for self-tuning optimizers to grow: hyperparameter tuning (or the lack thereof) is still responsible for a large share of an optimizer's practical performance.

\clearpage
\section*{Acknowledgements}
The authors would like to thank Google for providing a GCP grant that supported this work. We also extend our gratitude to the MLCommons organization and the members of the Algorithms working group for their ongoing support and collaboration. 

We are particularly grateful to Aaron Defazio for his intellectual contributions during working group discussions, which led to important fixes for the \algoperfnew benchmark. Finally, we thank Armstrong Foundjem for reviewing the manuscript and providing valuable feedback.

\bibliographystyle{plainnat}
\bibliography{bibliography}

\clearpage
\appendix
\crefalias{section}{appendix}
\section{Training Time Tables}
\label{app:step_execution_times}

\begin{table}[htbp]
  \centering
  \caption{Time to target as a fraction of each workload's self-tuning-ruleset time budget ($1.5\times$ the external-tuning maximum allowed runtime). \textemdash{} = target not reached.}
  \label{tab:time_to_target}
  \resizebox{\textwidth}{!}{%
  \begin{tabular}{lrrrrrrrrr}
    \toprule
    Submission & \criteo & \fastmri & \finewebedu & \resnet & \vit & \conformer & \deepspeech & \ogbg & \wmt \\
    \midrule
      \sfadamwii & 0.56 & 0.27 & \textemdash{} & \textemdash{} & \textemdash{} & 0.85 & 0.76 & 0.37 & 0.43 \\
      \ademamixadamw & 0.59 & 0.37 & 0.38 & \textemdash{} & \textemdash{} & 0.75 & \textemdash{} & 0.37 & 0.72 \\
      \sfadamw & 0.77 & 0.29 & \textemdash{} & \textemdash{} & \textemdash{} & \textemdash{} & 0.73 & 0.21 & 0.83 \\
      \nadamwbase & 0.61 & 0.39 & 0.43 & \textemdash{} & \textemdash{} & \textemdash{} & \textemdash{} & 0.24 & 0.62 \\
      \sfadamwjaxii & 0.61 & 0.45 & 0.32 & \textemdash{} & 0.88 & \textemdash{} & \textemdash{} & 0.32 & \textemdash{} \\
      \muonpt & 0.53 & \textemdash{} & 0.32 & \textemdash{} & \textemdash{} & \textemdash{} & \textemdash{} & 0.11 & 0.59 \\
      \sfadamwjax & \textemdash{} & 0.58 & 0.27 & \textemdash{} & 0.91 & \textemdash{} & \textemdash{} & 0.32 & 0.67 \\
      \nadamw & 0.56 & 0.44 & 0.54 & \textemdash{} & 0.75 & \textemdash{} & \textemdash{} & \textemdash{} & \textemdash{} \\
      \cautiousnadamw & \textemdash{} & 0.39 & 0.54 & \textemdash{} & \textemdash{} & \textemdash{} & \textemdash{} & 0.29 & 0.59 \\
      \muonjax & \textemdash{} & 0.45 & 0.43 & \textemdash{} & 0.80 & \textemdash{} & \textemdash{} & \textemdash{} & \textemdash{} \\
      \lion & \textemdash{} & 0.34 & 0.48 & \textemdash{} & \textemdash{} & \textemdash{} & \textemdash{} & 0.43 & 0.64 \\
      \ademamix & \textemdash{} & 0.44 & 0.27 & \textemdash{} & \textemdash{} & \textemdash{} & \textemdash{} & 0.29 & \textemdash{} \\
      \dilocosw & \textemdash{} & 0.63 & 0.59 & \textemdash{} & \textemdash{} & \textemdash{} & \textemdash{} & \textemdash{} & \textemdash{} \\
    \bottomrule
  \end{tabular}%
  }
\end{table}

\begin{table*}[htbp]
  \centering
  \caption{Step execution times, normalized to \nadamwbase. Each value is the ratio of a submission's average time per step to the reference's on that workload, excluding the first evaluation interval to remove compilation time; submissions run at their own batch sizes.}
  \label{tab:step_time_comparison}
  \resizebox{\textwidth}{!}{%
  \begin{tabular}{lrrrrrrrrr}
    \toprule
    Submission & \criteo & \fastmri & \finewebedu & \resnet & \vit & \conformer & \deepspeech & \ogbg & \wmt \\
    \midrule
    \multicolumn{10}{@{}l}{\textit{JAX submissions}} \\
    \midrule
      \sfadamwjax & $1.67$ & $1.25$ & $0.51$ & $1.04$ & $1.33$ & $1.00$ & $0.47$ & $1.05$ & $1.05$ \\
      \sfadamwjaxii & $1.00$ & $0.96$ & $0.52$ & $1.17$ & $1.10$ & $1.06$ & $1.04$ & $1.03$ & $1.05$ \\
      \muonjax & $1.96$ & $0.97$ & $1.16$ & $1.47$ & $1.24$ & $1.00$ & $1.00$ & $1.01$ & $1.39$ \\
      \nadamw & $1.01$ & $0.73$ & $1.00$ & $0.99$ & $1.00$ & $0.97$ & $0.91$ & $1.01$ & $1.01$ \\
      \nadamwbase & $1.00$ & $1.00$ & $1.00$ & $1.00$ & $1.00$ & $1.00$ & $1.00$ & $1.00$ & $1.00$ \\
      \cautiousnadamw & $1.03$ & $1.07$ & $1.01$ & $1.02$ & $1.08$ & $1.03$ & $1.03$ & $1.05$ & $1.00$ \\
      \dilocosw & $2.22$ & $1.15$ & $1.07$ & $1.78$ & $1.07$ & $1.02$ & $0.96$ & $1.02$ & $1.10$ \\
    \midrule
    \multicolumn{10}{@{}l}{\textit{PyTorch submissions}} \\
    \midrule
      \sfadamw & $1.38$ & $0.44$ & $0.90$ & $2.75$ & $1.69$ & $0.25$ & $0.16$ & $1.21$ & $1.18$ \\
      \sfadamwii & $0.99$ & $0.37$ & $0.91$ & $2.81$ & $1.87$ & $0.26$ & $0.19$ & $1.14$ & $1.12$ \\
      \ademamixadamw & $1.45$ & $0.60$ & $0.93$ & $2.40$ & $1.70$ & $0.28$ & $0.30$ & $1.20$ & $1.23$ \\
      \ademamix & $1.85$ & $0.81$ & $0.96$ & $2.52$ & $1.70$ & $0.28$ & $0.29$ & $1.20$ & $1.24$ \\
      \muonpt & $1.24$ & $0.46$ & $0.90$ & $2.56$ & $1.81$ & $0.29$ & $0.30$ & $1.24$ & $1.16$ \\
      \lion & $1.02$ & $0.43$ & $0.89$ & $2.41$ & $1.72$ & $0.28$ & $0.29$ & $1.13$ & $1.13$ \\
    \bottomrule
  \end{tabular}%
  }
\end{table*}

\clearpage
\section{Step-Based Leaderboard}
\label{app:step_based_leaderboard}
\begin{table}[htbp]
  \centering
  \caption{Wall-clock vs.\ step-based benchmark scores. Both use the same performance-profile scoring; only the notion of training time differs (seconds vs.\ optimizer steps to target). $\Delta$ is the rank change when moving from wall-clock to step-based scoring.}
  \label{tab:leaderboard_steps}
  \begin{tabular}{lrrrrc}
    \toprule
    & \multicolumn{2}{c}{Wall-clock} & \multicolumn{2}{c}{Steps} & \\
    \cmidrule(lr){2-3}\cmidrule(lr){4-5}
    Submission & Score & Rank & Score & Rank & $\Delta$ \\
    \midrule
    \sfadamwii & \textbf{0.5679} & 1 & \textbf{0.5253} & 1 & -- \\
    \ademamixadamw & 0.5191 & 2 & 0.5146 & 2 & -- \\
    \sfadamw & 0.4669 & 3 & 0.4461 & 3 & -- \\
    \nadamwbase & 0.4506 & 4 & 0.4422 & 4 & -- \\
    \sfadamwjaxii & 0.4405 & 5 & 0.3945 & 6 & $\downarrow$1 \\
    \muonpt & 0.4231 & 6 & 0.4220 & 5 & $\uparrow$1 \\
    \sfadamwjax & 0.4108 & 7 & 0.3945 & 6 & $\uparrow$1 \\
    \nadamw & 0.3844 & 8 & 0.3883 & 8 & -- \\
    \cautiousnadamw & 0.3143 & 9 & 0.3042 & 10 & $\downarrow$1 \\
    \muonjax & 0.2845 & 10 & 0.3098 & 9 & $\uparrow$1 \\
    \lion & 0.2783 & 11 & 0.2576 & 12 & $\downarrow$1 \\
    \ademamix & 0.2464 & 12 & 0.2637 & 11 & $\uparrow$1 \\
    \dilocosw & 0.1302 & 13 & 0.1672 & 13 & -- \\
    \bottomrule
  \end{tabular}
\end{table}

\leaderboardprofilefigure
    {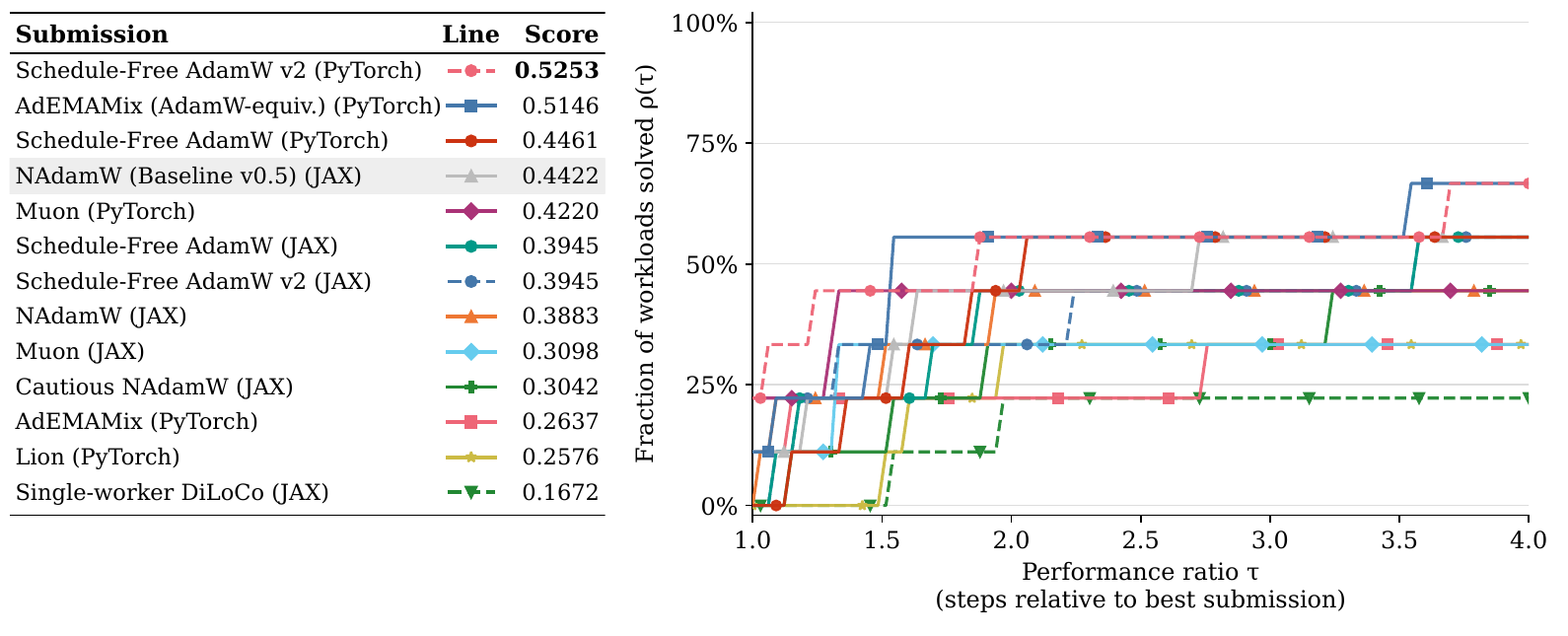}
    {AlgoPerf self-tuning leaderboard and performance profiles (steps to target).}
    {fig:step_leaderboard_profiles}
    {tab:step_count_scores}
    {fig:step_count_profiles}

\begin{table}[htbp]
  \centering
  \caption{Median number of optimizer steps to reach the validation target. \textemdash{} = target not reached. Submissions choose their own batch sizes, so step counts compare optimizer updates, not examples seen.}
  \label{tab:steps_to_target}
  \resizebox{\textwidth}{!}{%
  \begin{tabular}{lrrrrrrrrr}
    \toprule
    Submission & \criteo & \fastmri & \finewebedu & \resnet & \vit & \conformer & \deepspeech & \ogbg & \wmt \\
    \midrule
      \sfadamwii & 7,951 & 12,431 & \textemdash{} & \textemdash{} & \textemdash{} & 83,184 & 78,669 & 28,449 & 73,222 \\
      \ademamixadamw & 8,371 & 10,206 & 44,503 & \textemdash{} & \textemdash{} & 67,706 & \textemdash{} & 27,247 & 112,742 \\
      \sfadamw & 10,346 & 10,599 & \textemdash{} & \textemdash{} & \textemdash{} & \textemdash{} & 90,588 & 15,853 & 133,517 \\
      \nadamwbase & 9,145 & 6,648 & 47,039 & \textemdash{} & \textemdash{} & \textemdash{} & \textemdash{} & 20,879 & 118,600 \\
      \muonpt & 7,716 & \textemdash{} & 39,220 & \textemdash{} & \textemdash{} & \textemdash{} & \textemdash{} & 7,711 & 97,224 \\
      \sfadamwjaxii & 9,005 & 8,750 & 68,640 & \textemdash{} & 164,206 & \textemdash{} & \textemdash{} & 27,457 & \textemdash{} \\
      \sfadamwjax & \textemdash{} & 7,160 & 57,242 & \textemdash{} & 175,889 & \textemdash{} & \textemdash{} & 27,341 & 122,198 \\
      \nadamw & 8,286 & 9,924 & 58,784 & \textemdash{} & 152,461 & \textemdash{} & \textemdash{} & \textemdash{} & \textemdash{} \\
      \muonjax & \textemdash{} & 8,801 & 40,724 & \textemdash{} & 151,926 & \textemdash{} & \textemdash{} & \textemdash{} & \textemdash{} \\
      \cautiousnadamw & \textemdash{} & 7,518 & 58,199 & \textemdash{} & \textemdash{} & \textemdash{} & \textemdash{} & 24,948 & 112,372 \\
      \ademamix & \textemdash{} & 7,507 & 30,636 & \textemdash{} & \textemdash{} & \textemdash{} & \textemdash{} & 21,236 & \textemdash{} \\
      \lion & \textemdash{} & 10,645 & 59,524 & \textemdash{} & \textemdash{} & \textemdash{} & \textemdash{} & 33,386 & 109,080 \\
      \dilocosw & \textemdash{} & 10,140 & 59,550 & \textemdash{} & \textemdash{} & \textemdash{} & \textemdash{} & \textemdash{} & \textemdash{} \\
    \bottomrule
  \end{tabular}%
  }
\end{table}

\clearpage
\section{Relaxed Leaderboard}
\label{app:relaxed_leaderboard}

\begin{table}[htbp]
  \centering
  \caption{Standard vs.\ 10\% relaxed-target AlgoPerf self-tuning leaderboard. $\Delta$ is the rank change after relaxing targets.}
  \label{tab:scores_relaxed}
  \begin{tabular}{lrrrrc}
    \toprule
    & \multicolumn{2}{c}{Standard} & \multicolumn{2}{c}{10\% relaxed} & \\
    \cmidrule(lr){2-3}\cmidrule(lr){4-5}
    Submission & Score & Rank & Score & Rank & $\Delta$ \\
    \midrule
    \sfadamwjaxii & 0.4405 & 5 & \textbf{0.6852} & 1 & $\uparrow$4 \\
    \ademamix & 0.2464 & 12 & 0.6246 & 2 & $\uparrow$10 \\
    \sfadamwii & \textbf{0.5679} & 1 & 0.6072 & 3 & $\downarrow$2 \\
    \muonpt & 0.4231 & 6 & 0.5993 & 4 & $\uparrow$2 \\
    \sfadamw & 0.4669 & 3 & 0.5786 & 5 & $\downarrow$2 \\
    \sfadamwjax & 0.4108 & 7 & 0.5561 & 6 & $\uparrow$1 \\
    \ademamixadamw & 0.5191 & 2 & 0.5213 & 7 & $\downarrow$5 \\
    \nadamwbase & 0.4506 & 4 & 0.4933 & 8 & $\downarrow$4 \\
    \muonjax & 0.2845 & 10 & 0.4506 & 9 & $\uparrow$1 \\
    \lion & 0.2783 & 11 & 0.4439 & 10 & $\uparrow$1 \\
    \cautiousnadamw & 0.3143 & 9 & 0.4012 & 11 & $\downarrow$2 \\
    \nadamw & 0.3844 & 8 & 0.3754 & 12 & $\downarrow$4 \\
    \dilocosw & 0.1302 & 13 & 0.2318 & 13 & -- \\
    \bottomrule
  \end{tabular}
\end{table}

\leaderboardprofilefigure
    {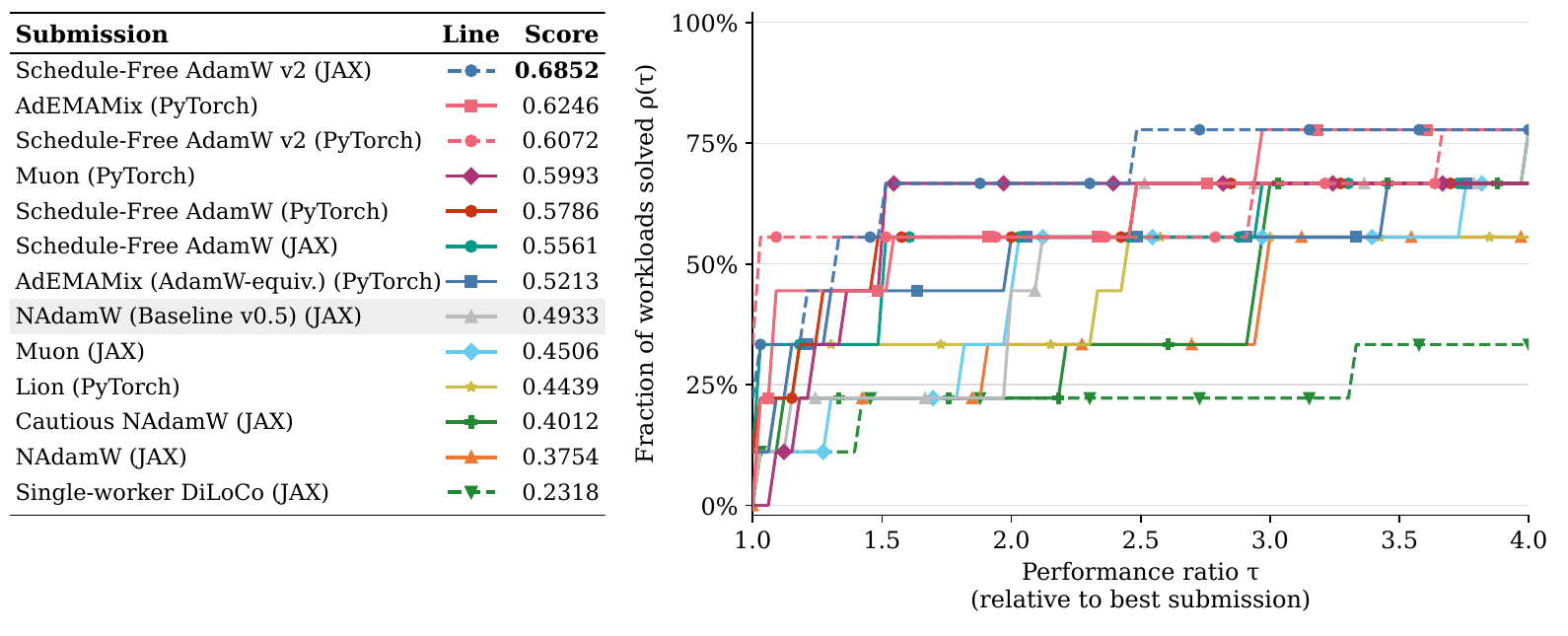}
    {AlgoPerf self-tuning leaderboard and performance profiles with $10\%$ relaxed targets.}
    {fig:relaxed_leaderboard_profiles}
    {tab:relaxed_leaderboard_scores}
    {fig:performance_profiles_relaxed}

\begin{table}[htbp]
  \centering
  \caption{Time to target with targets relaxed 10\%, as a fraction of each workload's self-tuning-ruleset time budget ($1.5\times$ the external-tuning maximum allowed runtime). \textbf{Bold} values mark targets reached only after relaxation; \textemdash{} = target not reached.}
  \label{tab:time_to_target_relaxed}
  \resizebox{\textwidth}{!}{%
  \begin{tabular}{lrrrrrrrrr}
    \toprule
    Submission & \criteo & \fastmri & \finewebedu & \resnet & \vit & \conformer & \deepspeech & \ogbg & \wmt \\
    \midrule
      \sfadamwjaxii & 0.03 & 0.07 & 0.21 & \textbf{0.11} & 0.27 & \textemdash{} & \textemdash{} & 0.13 & \textbf{0.08} \\
      \ademamix & \textbf{0.03} & 0.09 & 0.16 & \textemdash{} & \textemdash{} & \textbf{0.67} & \textbf{0.44} & 0.16 & \textbf{0.14} \\
      \sfadamwii & 0.03 & 0.06 & \textbf{0.59} & \textemdash{} & \textemdash{} & 0.61 & 0.42 & 0.16 & 0.06 \\
      \muonpt & 0.03 & \textbf{0.08} & 0.22 & \textemdash{} & \textemdash{} & \textbf{0.72} & \textemdash{} & 0.08 & 0.08 \\
      \sfadamw & 0.03 & 0.08 & \textemdash{} & \textemdash{} & \textemdash{} & \textbf{0.72} & 0.42 & 0.13 & 0.08 \\
      \sfadamwjax & \textbf{0.03} & 0.09 & 0.16 & \textemdash{} & 0.27 & \textemdash{} & \textemdash{} & 0.16 & 0.08 \\
      \ademamixadamw & 0.03 & 0.08 & 0.32 & \textemdash{} & \textemdash{} & 0.67 & \textbf{0.47} & 0.19 & 0.24 \\
      \nadamwbase & 0.03 & 0.07 & 0.32 & \textbf{0.43} & \textbf{0.56} & \textemdash{} & \textemdash{} & 0.13 & 0.11 \\
      \muonjax & \textbf{0.05} & 0.08 & 0.32 & \textbf{0.40} & 0.48 & \textemdash{} & \textemdash{} & \textbf{0.05} & \textbf{0.27} \\
      \lion & \textbf{0.03} & 0.07 & 0.38 & \textemdash{} & \textemdash{} & \textemdash{} & \textbf{0.44} & 0.27 & 0.13 \\
      \cautiousnadamw & \textbf{0.03} & 0.07 & 0.48 & \textbf{0.43} & \textbf{0.59} & \textemdash{} & \textemdash{} & 0.16 & 0.16 \\
      \nadamw & 0.03 & 0.06 & 0.48 & \textbf{0.45} & 0.51 & \textemdash{} & \textemdash{} & \textbf{0.16} & \textbf{0.27} \\
      \dilocosw & \textbf{0.03} & 0.09 & 0.54 & \textbf{0.61} & \textemdash{} & \textemdash{} & \textemdash{} & \textbf{0.27} & \textbf{0.27} \\
    \bottomrule
  \end{tabular}%
  }
\end{table}

\clearpage
\section{Training Curves}
\label{sec:training_curves}

\begin{figure}[htbp]
    \centering
    \includegraphics[width=\linewidth]{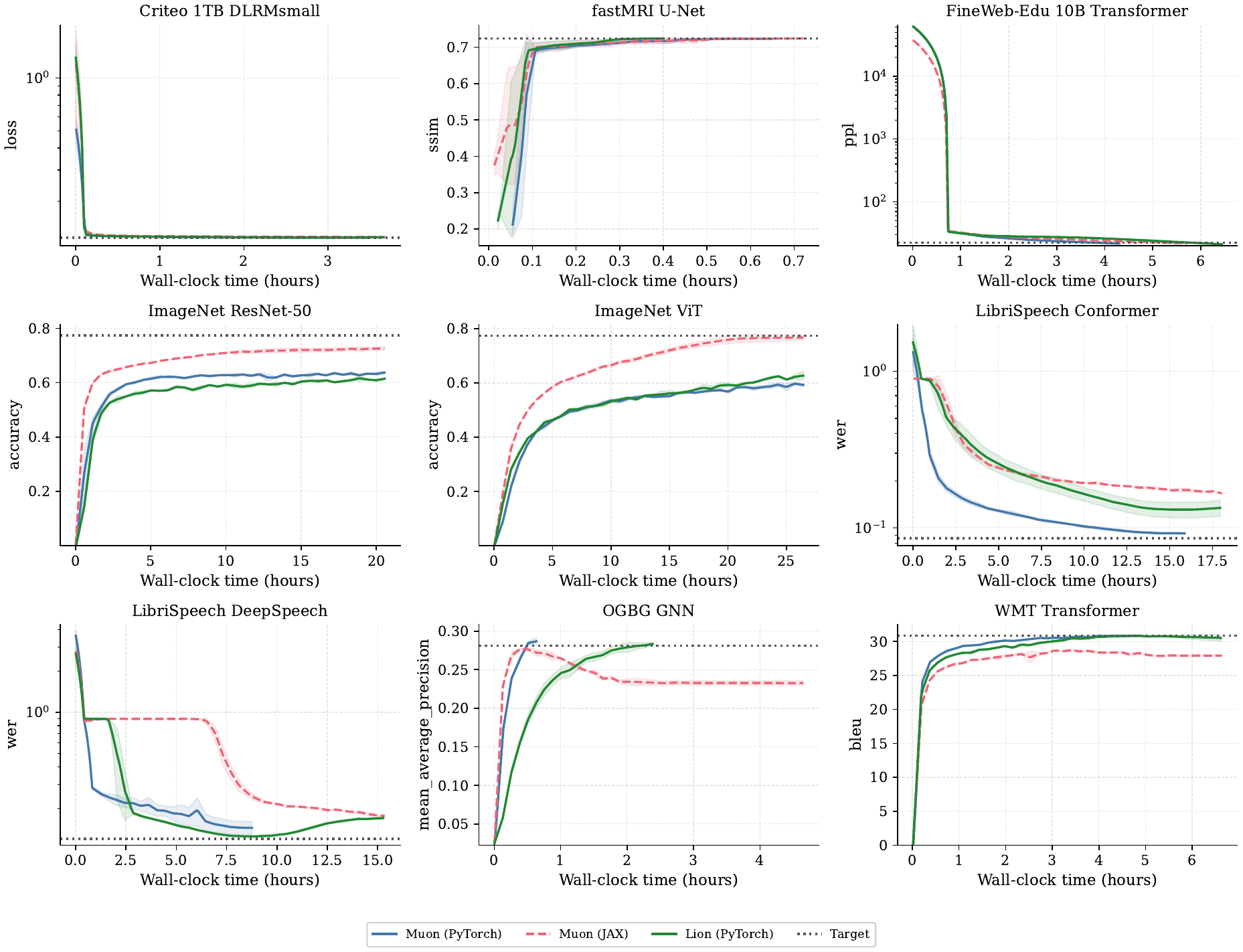}
    \caption{Validation metric against wall-clock time for \muonpt{}, \muonjax{}, and \lion{} on the nine base workloads. Lines show the mean over three studies and shaded bands show $\pm 1$ standard deviation. The dotted line marks the validation target. Solid lines are PyTorch and dashed lines are JAX implementations. Minimized metrics (loss, WER, perplexity) are on a logarithmic axis. Each curve ends when its longest study stops, either at the target or at the time budget. \muonjax{} uses different fixed hyperparameters from \muonpt{}. The two differ most on ImageNet ResNet-50 and ImageNet ViT, where \muonjax{} reaches higher accuracy, and on OGBG GNN and WMT Transformer, where it levels off below the target.}
    \label{fig:muon_lion_training_curves}
\end{figure}

\begin{figure}[htbp]
    \centering
    \includegraphics[width=\linewidth]{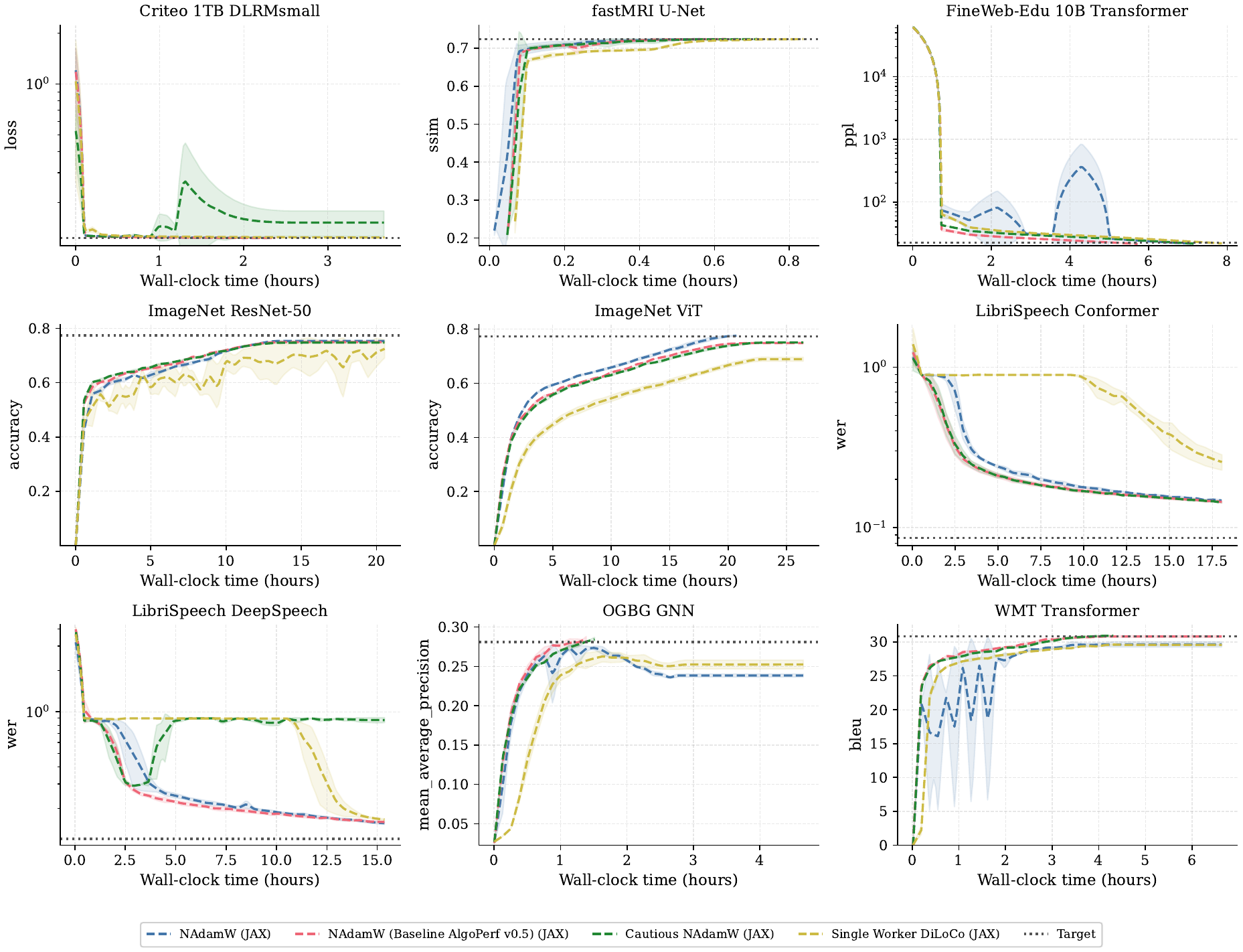}
    \caption{Validation metric against wall-clock time for \nadamw{}, \nadamwbase{}, \cautiousnadamw{}, and \dilocosw{} on the nine base workloads. All four are JAX implementations. Lines show the mean over three studies and shaded bands show $\pm 1$ standard deviation. Minimized metrics (loss, WER, perplexity) are on a logarithmic axis. \dilocosw{} trails the NAdamW variants on ImageNet ViT, LibriSpeech Conformer, and OGBG GNN. \cautiousnadamw{} shows a loss spike on Criteo 1TB DLRMsmall and regresses after about four hours on LibriSpeech DeepSpeech.}
    \label{fig:nadamw_diloco_training_curves}
\end{figure}

\begin{figure}[htbp]
    \centering
    \includegraphics[width=\linewidth]{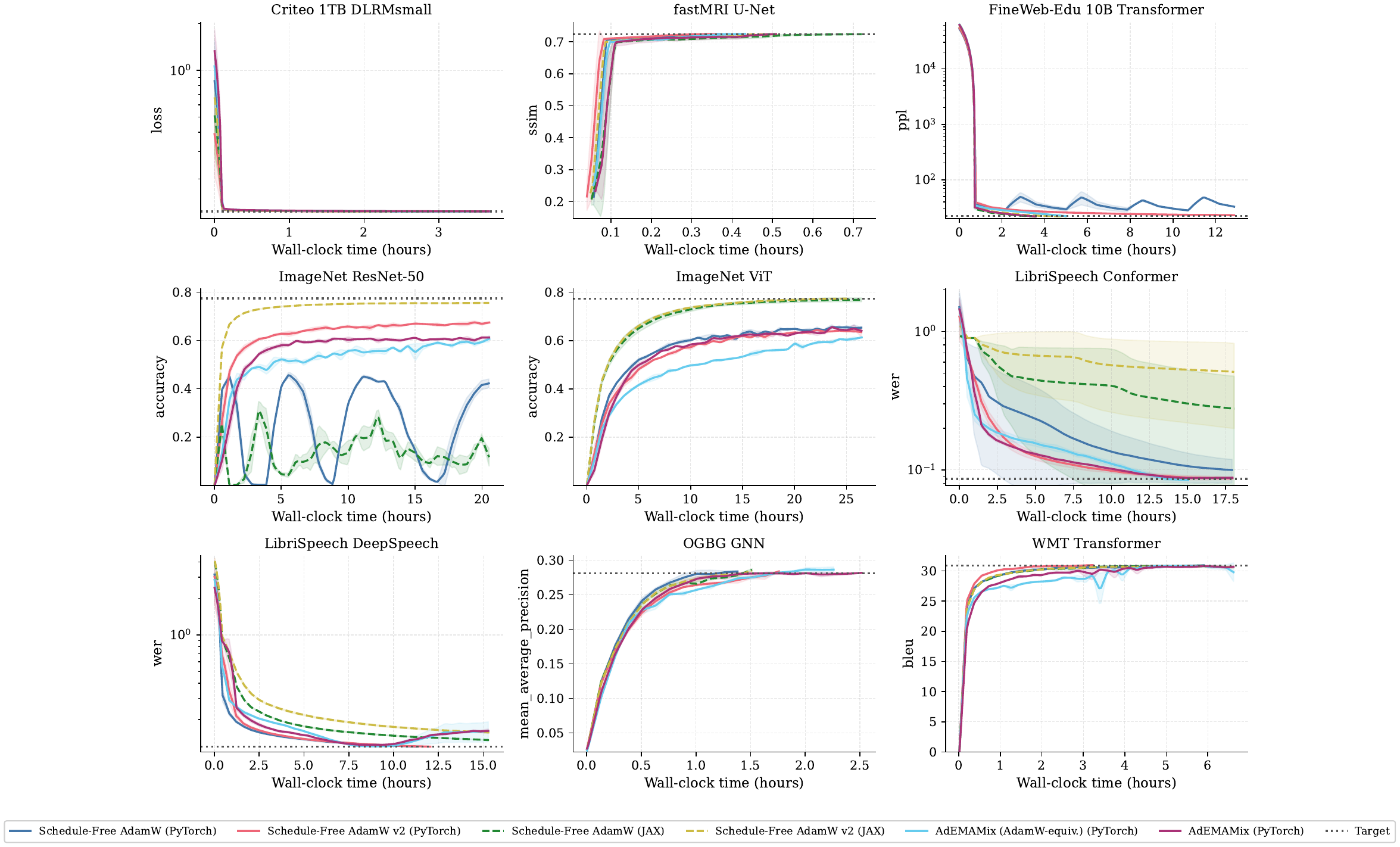}
    \caption{Validation metric against wall-clock time for the four Schedule-Free AdamW submissions (\sfadamw{}, \sfadamwii{}, \sfadamwjax{}, \sfadamwjaxii{}) and the two AdEMAMix submissions (\ademamixadamw{}, \ademamix{}) on the nine base workloads. Lines show the mean over three studies and shaded bands show $\pm 1$ standard deviation. The dotted line marks the validation target. Solid lines are PyTorch and dashed lines are JAX implementations. Minimized metrics (loss, WER, perplexity) are on a logarithmic axis. On ImageNet ResNet-50, the first versions of Schedule-Free AdamW (\sfadamw{}, \sfadamwjax{}) oscillate while both v2 submissions train stably. On ImageNet ViT, the JAX implementations nearly reach the target while the PyTorch submissions level off near 0.65 accuracy.}
    \label{fig:sfadamw_ademamix}
\end{figure}

\clearpage
\section{Hyperparameter Sweep Details and Search Spaces}
\label{sec:appendix_sweep_details}

For our hyperparameter sensitivity and retention analysis (\cref{sec:tuning}), we ran sweeps of $N = 100$ configurations across all nine \algoperf workloads for NAdamW, Single-Worker DiLoCo, Muon, and AdEMAMix. For each optimizer, configurations were sampled using quasi-random Halton sequences. The same 100 configurations were evaluated across all nine workloads, allowing direct comparison between universal ($\bar{S}(c)$) and per-workload ($S_w(c)$) selection. All runs used linear learning rate warmup followed by cosine decay to zero at step budget $N_{\max, w}$, with gradient clipping set to $1.0$ and $\epsilon = 10^{-8}$ where applicable. \cref{tab:appendix_search_spaces} lists the search spaces and fixed parameters.

\begin{table}[htbp]
\centering
\caption{Hyperparameter search space specifications for the $N=100$ quasi-random Halton sweeps evaluated across all nine \algoperf workloads.}
\label{tab:appendix_search_spaces}
\resizebox{\linewidth}{!}{%
\begin{tabular}{llcll}
\toprule
Optimizer & Tuned Hyperparameter & Sampling Range / Values & Distribution & Fixed Parameters \& Architectural Notes \\
\midrule
\multirow{7}{*}{\textbf{NAdamW} ($D=7$)} 
  & Learning rate ($\eta$) & $[10^{-5},\, 5 \times 10^{-3}]$ & Log-uniform & Cosine decay schedule (\texttt{cosine\_warmup}) \\
  & Momentum $1 - \beta_1$ & $[10^{-3},\, 0.5]$ & Log-uniform & Fixed $\epsilon = 10^{-8}$, bias correction (\texttt{debias = True}) \\
  & Second moment $1 - \beta_2$ & $[10^{-3},\, 0.5]$ & Log-uniform & Tied dropout rates (\texttt{tie\_dropouts = True}, when supported) \\
  & Weight decay ($\lambda$) & $[10^{-2},\, 0.5]$ & Log-uniform & \texttt{l2\_decay\_factor = None} \\
  & LR warmup fraction ($T_{\text{warmup}}$) & $\{0.02,\, 0.05,\, 0.10\}$ & Discrete uniform & $^*$Tuned only on workloads supporting the regularizer \\
  & Label smoothing$^*$ & $\{0.0,\, 0.2\}$ & Discrete uniform & \\
  & Dropout rate$^*$ & $\{0.0,\, 0.1\}$ & Discrete uniform & \\
\midrule
\multirow{5}{*}{\textbf{Schedule-Free AdamW} ($D=5$)} 
  & Learning rate ($\eta$) & $[10^{-4},\, 10^{-1}]$ & Log-uniform & Polynomial iterate averaging power \texttt{weight\_lr\_power} $= 2.0$ \\
  & Momentum $1 - \beta_1$ & $[10^{-4},\, 0.5]$ & Log-uniform & Evaluates averaged parameters (\texttt{eval\_use\_ema = True}) \\
  & Second moment $1 - \beta_2$ & $[10^{-4},\, 0.5]$ & Log-uniform & Fixed $\epsilon = 10^{-8}$, \texttt{cosine\_warmup} schedule \\
  & Weight decay ($\lambda$) & $[10^{-6},\, 1.0]$ & Log-uniform & Fixed regularizers (when supported): \texttt{dropout\_rate} $= 0.1$, \\
  & LR warmup fraction ($T_{\text{warmup}}$) & $\{0.02,\, 0.05,\, 0.10\}$ & Discrete uniform & \quad \texttt{aux\_dropout\_rate} $= 0.1$, \texttt{label\_smoothing} $= 0.1$ \\
\midrule
\multirow{5}{*}{\textbf{Single-Worker DiLoCo} ($D=5$)} 
  & Inner learning rate ($\eta_{\text{inner}}$) & $[3 \times 10^{-4},\, 10^{-3}]$ & Log-uniform & Inner AdamW: fixed $\beta_1 = 0.99$, $\beta_2 = 0.995$, $\epsilon = 10^{-8}$ \\
  & Inner weight decay ($\lambda$) & $[0.1,\, 0.4]$ & Uniform & Fixed LR warmup fraction: $0.0$ (\texttt{cosine\_warmup} schedule) \\
  & Synchronization period ($H$) & $\{50,\, 75,\, 150\}$ & Discrete uniform & Fixed regularizers (when supported): \texttt{dropout\_rate} $= 0.0$, \\
  & Outer learning rate ($\eta_{\text{outer}}$) & $[0.2,\, 0.4]$ & Uniform & \quad \texttt{tie\_dropouts = True} \\
  & Outer momentum ($\mu_{\text{outer}}$) & $[0.75,\, 0.85]$ & Uniform & Single-worker outer Nesterov step every $H$ inner steps \\
\midrule
\multirow{9}{*}{\textbf{Muon} ($D=9$)} 
  & Base learning rate ($\eta$) & $[10^{-5},\, 10^{-2}]$ & Log-uniform & 2D matrix parameters optimized via Newton-Schulz \\
  & Matrix momentum ($\beta_{\text{muon}}$) & $[0.2,\, 0.999]$ & Uniform & Non-matrix (1D/embedding) parameters via paired NAdamW \\
  & Matrix LR multiplier ($m_{\text{lr}}$) & $[10^{-2},\, 10^{4}]$ & Log-uniform & Paired NAdamW: fixed $\epsilon = 10^{-8}$, \texttt{debias = True} \\
  & Matrix weight decay ($\lambda_{\text{muon}}$) & $[10^{-3},\, 0.5]$ & Log-uniform & Cosine decay schedule (\texttt{cosine\_warmup}) \\
  & Matrix Nesterov flag & $\{\text{True},\, \text{False}\}$ & Discrete uniform & Fixed regularizers (when supported): \texttt{dropout\_rate} $= 0.1$, \\
  & Paired 1D momentum $1 - \beta_{1,\text{pair}}$ & $[10^{-5},\, 10^{-2}]$ & Log-uniform & \quad \texttt{aux\_dropout\_rate} $= 0.1$, \texttt{label\_smoothing} $= 0.1$, \\
  & Paired 1D second moment $1 - \beta_{2,\text{pair}}$ & $[10^{-3},\, 0.5]$ & Log-uniform & \quad \texttt{tie\_dropouts = True} \\
  & Paired 1D weight decay ($\lambda_{\text{pair}}$) & $[10^{-2},\, 0.5]$ & Log-uniform & \\
  & LR warmup fraction ($T_{\text{warmup}}$) & $\{0.02,\, 0.05,\, 0.10\}$ & Discrete uniform & \\
\midrule
\multirow{7}{*}{\textbf{AdEMAMix} ($D=7$)} 
  & Learning rate ($\eta$) & $[10^{-5},\, 10^{-1}]$ & Log-uniform & Dual-EMA momentum mixture ($m_1$ fast, $m_2$ slow, $\nu$ second moment) \\
  & Fast momentum $1 - \beta_1$ & $[10^{-4},\, 1.0]$ & Log-uniform & Fixed LR warmup fraction: $0.02$ (\texttt{cosine\_warmup} schedule) \\
  & Second moment $1 - \beta_2$ & $[10^{-4},\, 10^{-1}]$ & Log-uniform & Slow EMA scheduler warms up $\beta_3$ and $\alpha$ over $T_{\alpha,\beta_3}$ \\
  & Slow momentum $1 - \beta_3$ & $[10^{-5},\, 10^{-2}]$ & Log-uniform & Fixed $\epsilon = 10^{-8}$, gradient clipping \texttt{grad\_clip} $= 0.5$ \\
  & Slow mixture weight ($\alpha$) & $[1.0,\, 10.0]$ & Uniform & Fixed \texttt{dropout\_rate} $= 0.1$ (when supported) \\
  & Momentum warmup fraction ($T_{\alpha,\beta_3}$) & $[0.05,\, 1.0]$ & Uniform & \\
  & Weight decay ($\lambda$) & $[10^{-4},\, 1.0]$ & Log-uniform & \\
\bottomrule
\end{tabular}%
}
\end{table}

\clearpage
\section{Top-K Distributions}
\label{sec:topk_distributions}

\begin{figure}[htbp]
    \centering
    \includegraphics[width=\linewidth]        {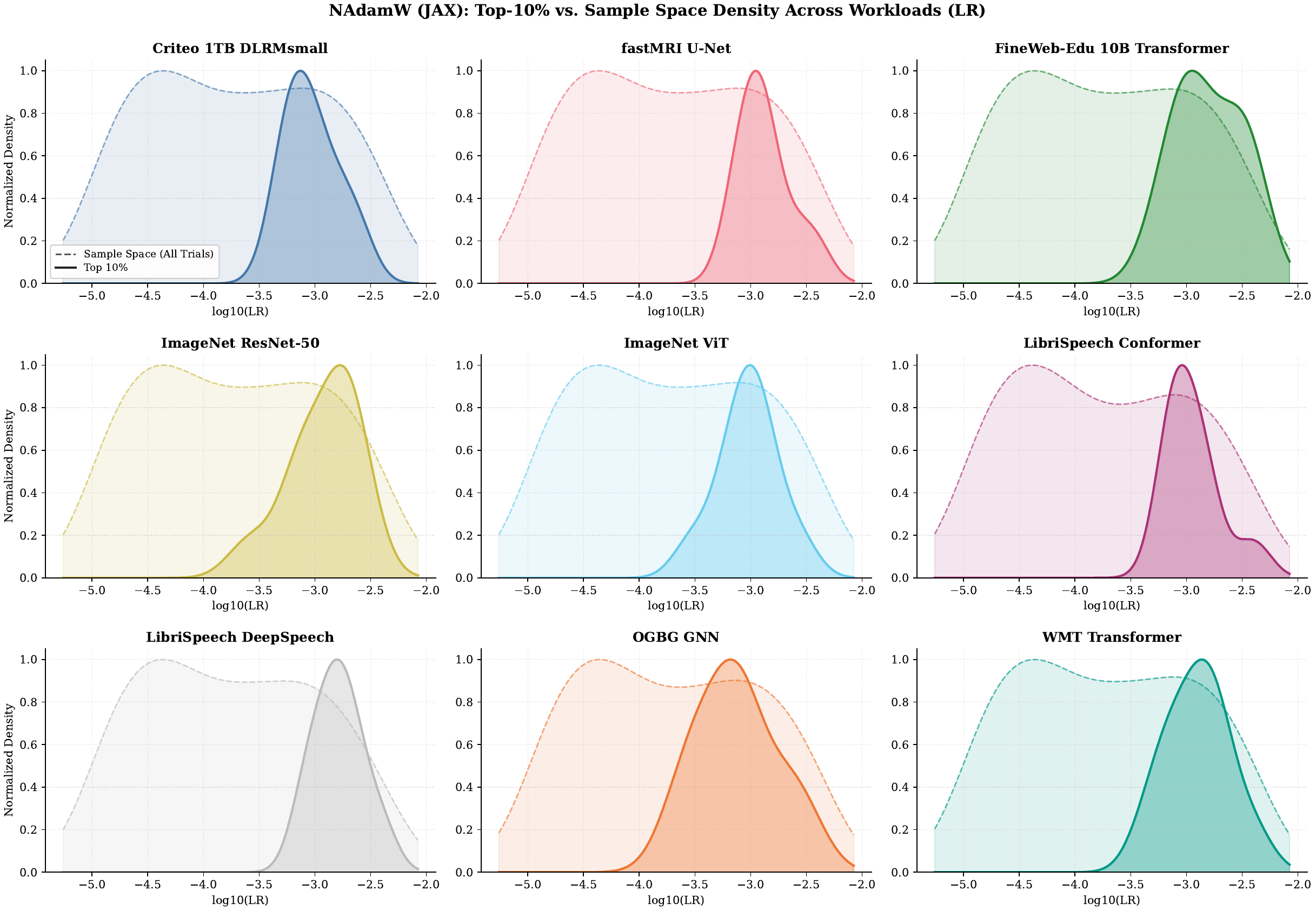}
    \caption{Comparison of sampled search space vs. optimal hyperparameter distributions. Empirical density (KDE) of hyperparameter values across workloads, comparing all sampled configurations (dashed line, light fill) to the top-10\% best-performing trials (solid line, dark fill). Regions where the top-10\% density is significantly more concentrated or peaked relative to the sample space indicate favorable hyperparameter settings for that workload.}
    \label{fig:topk_dist_nadamw}
\end{figure}

\begin{figure}[htbp]
    \centering
    \includegraphics[width=\linewidth]        {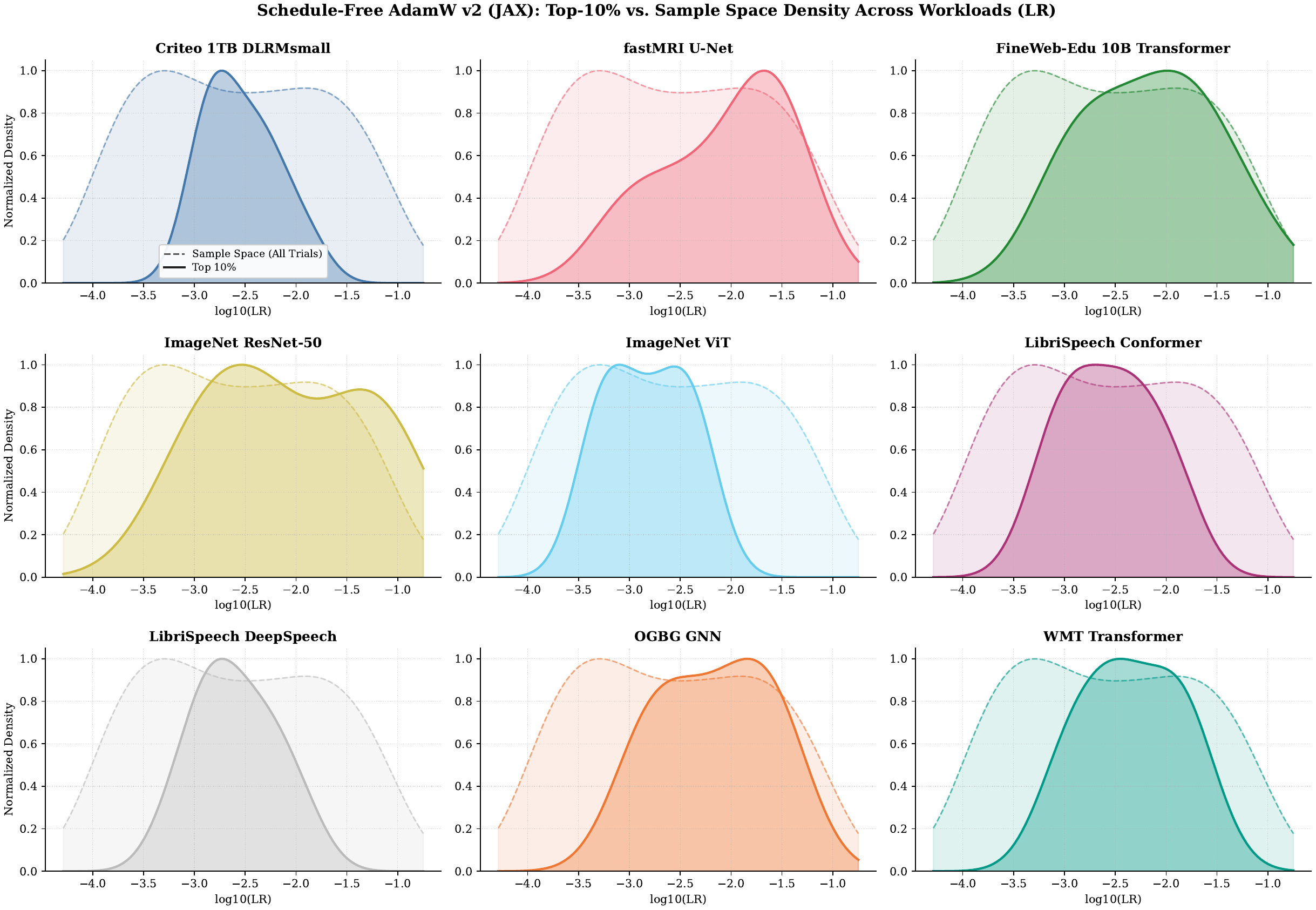}
    \caption{Comparison of sampled search space vs. optimal hyperparameter distributions. Empirical density (KDE) of hyperparameter values across workloads, comparing all sampled configurations (dashed line, light fill) to the top-10\% best-performing trials (solid line, dark fill). Regions where the top-10\% density is significantly more concentrated or peaked relative to the sample space indicate favorable hyperparameter settings for that workload.}
    \label{fig:topk_dist_sf}
\end{figure}

\begin{figure}[htbp]
    \centering
    \includegraphics[width=\linewidth]        {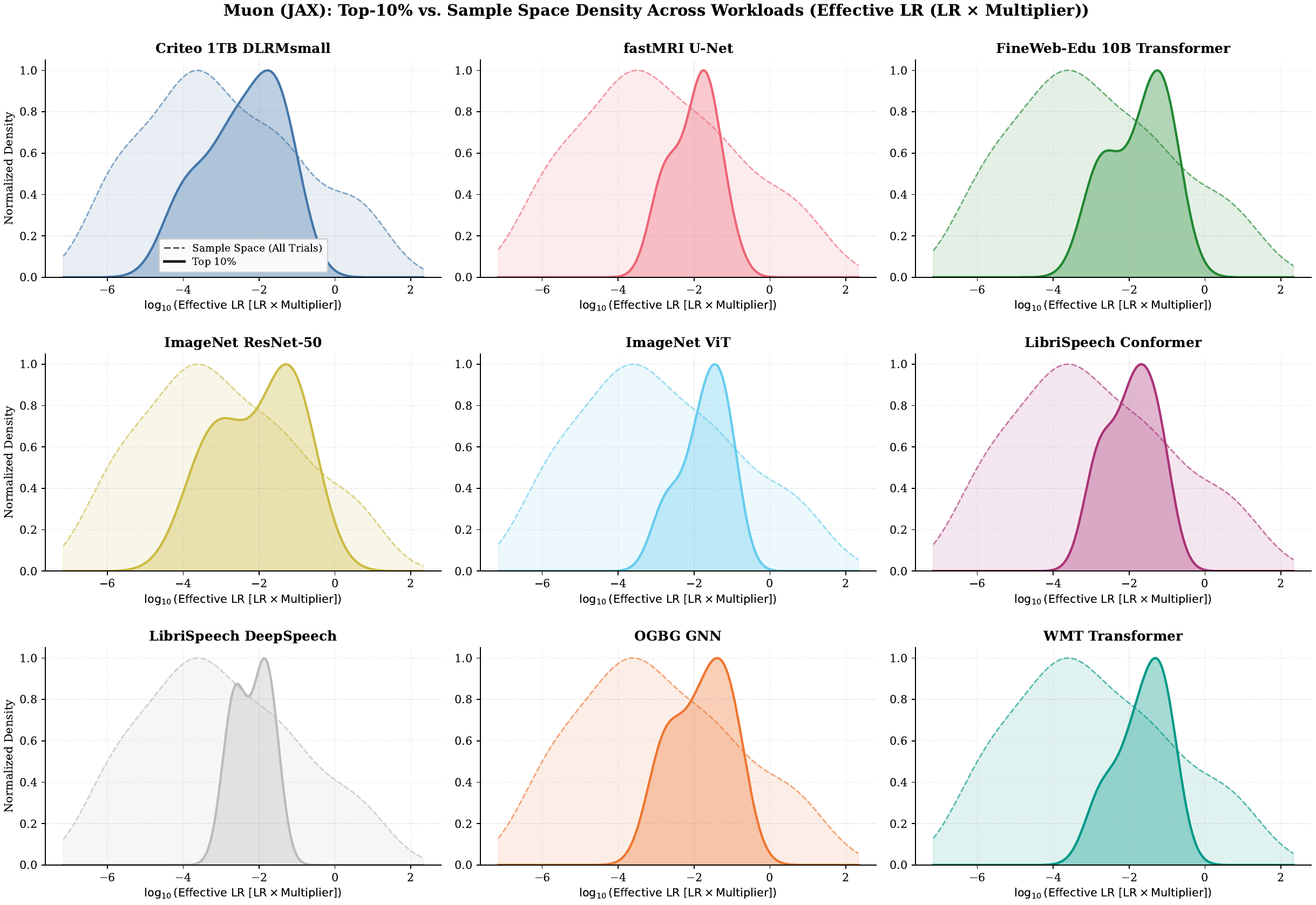}
    \caption{Comparison of sampled search space vs. optimal hyperparameter distributions. Empirical density (KDE) of hyperparameter values across workloads, comparing all sampled configurations (dashed line, light fill) to the top-10\% best-performing trials (solid line, dark fill). Regions where the top-10\% density is significantly more concentrated or peaked relative to the sample space indicate favorable hyperparameter settings for that workload.}
    \label{fig:topk_dist_muon}
\end{figure}

\begin{figure}[htbp]
    \centering
    \includegraphics[width=\linewidth]        {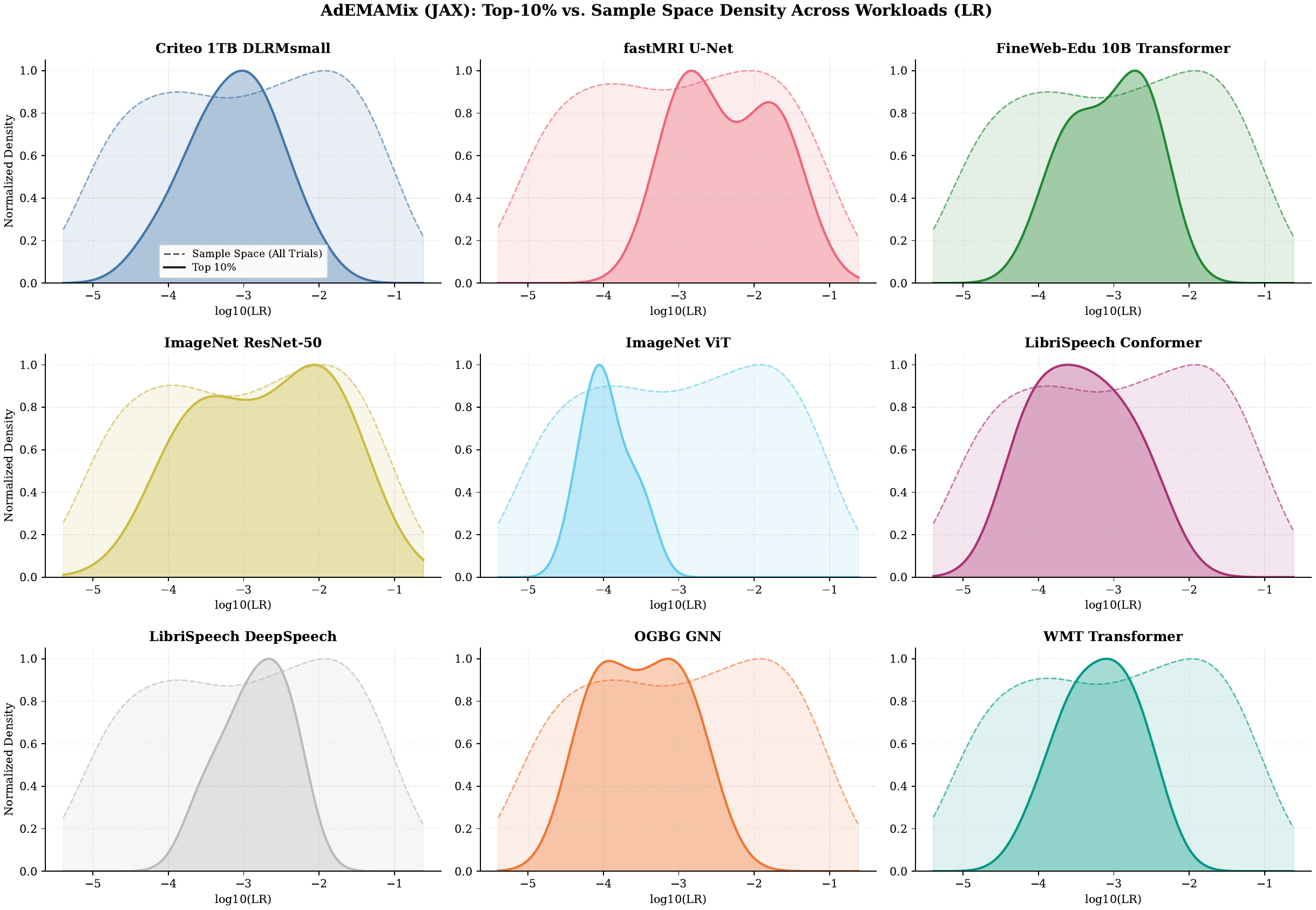}
    \caption{Comparison of sampled search space vs. optimal hyperparameter distributions. Empirical density (KDE) of hyperparameter values across workloads, comparing all sampled configurations (dashed line, light fill) to the top-10\% best-performing trials (solid line, dark fill). Regions where the top-10\% density is significantly more concentrated or peaked relative to the sample space indicate favorable hyperparameter settings for that workload.}
    \label{fig:topk_dist_ademamix}
\end{figure}

\Needspace{14\baselineskip}
\section{Efficiency and Systems Analysis}
\label{sec:systems_analysis}
\subsection{JAX vs.\ PyTorch}
\label{sec:framework_comparison}

Similarly to the engineering challenges presented in \algoperfold~\citep{Kasimbeg2025AlgoPerfResults}, \algoperfnew also suffers from a disparity in performance between the two neural network frameworks, JAX
\& PyTorch. We analyze this discrepancy by comparing Schedule-Free AdamW (v1 and v2), which were submitted in both frameworks and share their core optimizer hyperparameters.

\begin{figure}[htbp]
    \centering
    \includegraphics[width=\textwidth]{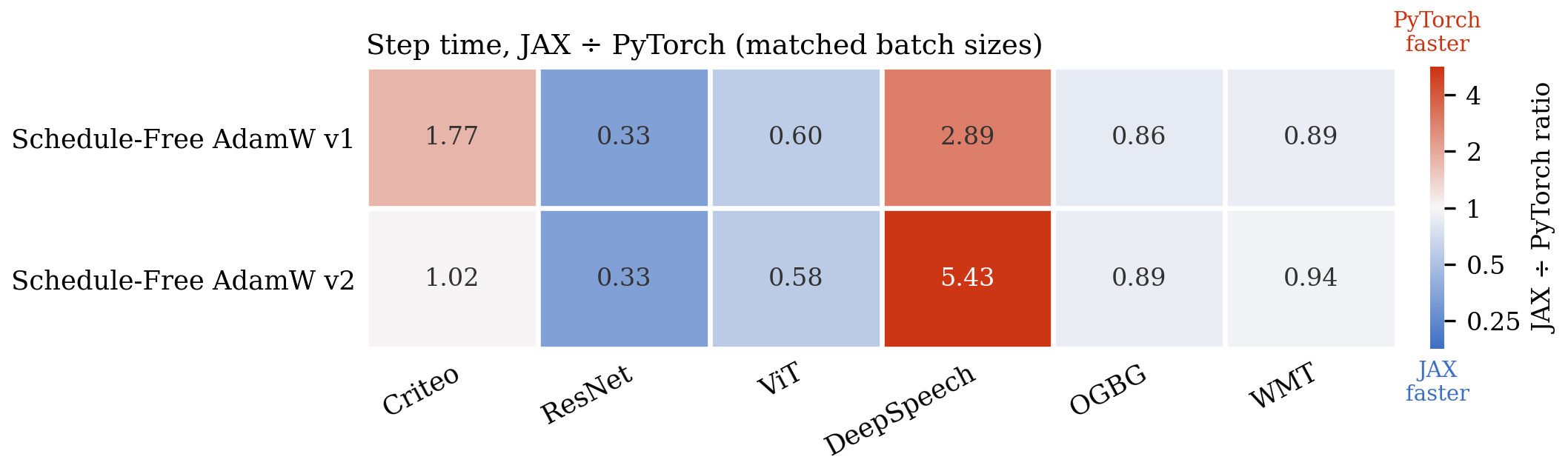}
    \caption{JAX vs.\ PyTorch median time per training step. Batch sizes and hyperparameters are held constant across frameworks.}
    \label{fig:framework_comparison}
\end{figure}

\cref{fig:framework_comparison} compares each pair on median time per training step. We exclude fastMRI, FineWeb-Edu, and Librispeech Conformer due to their differences in training batch sizes.

We see that there are still clear step time discrepancies varying by the workload. DeepSpeech is significantly advantaged toward PyTorch, whereas ResNet and ViT are faster in JAX by a large margin. The DeepSpeech disparity was already known in
\algoperfold~\citep{Kasimbeg2025AlgoPerfResults}. However, we disabled the faster JAX cuDNN LSTM implementation, which more closely matches PyTorch, because of a gradient-calculation bug.
Why ResNet and ViT still appear significantly JAX-biased is an ongoing investigation.

\end{document}